\documentclass{article}

\usepackage{arxiv}

\usepackage[utf8]{inputenc} 
\usepackage[T1]{fontenc}    
\usepackage{hyperref}       
\usepackage{url}            
\usepackage{booktabs}       
\usepackage{amsfonts}       
\usepackage{nicefrac}       
\usepackage{microtype}      
\usepackage{lipsum}
\usepackage{graphicx}
\usepackage{algorithm}
\usepackage{algorithmic}
\usepackage{graphicx}
\usepackage{textcomp}
\usepackage{multirow}
\usepackage{caption}
\usepackage{amsmath,amssymb,amsfonts}
\usepackage{hyperref}
\graphicspath{ {./images/} }
\usepackage[numbers,sort&compress]{natbib}  

\title{TWLR: Text-Guided Weakly-Supervised Lesion Localization and Severity Regression for Explainable Diabetic Retinopathy Grading}

\author{
Xi Luo$^{1,2,3}$, Shixin Xu$^{4}$, Ying Xie$^{5}$, JianZhong Hu$^{6}$, Yuwei He$^{1,2,3}$,  Yuhui Deng$^{1,2,3,*}$, Huaxiong Huang$^{4,7,8}\thanks{Correspondence:\href{mailto:ivandeng@bnbu.edu.cn}{ivandeng@bnbu.edu.cn} or \href{mailto:hhuang@bnbu.edu.cn}{hhuang@bnbu.edu.cn}}$\\
\small
$^1$Guangdong Provincial Key Laboratory of Interdisciplinary Research and Application for Data Science\\
$^2$Department of Statistics and Data Science, Beijing Normal-Hong Kong Baptist University \\
$^3$Faculty of Science, Hong Kong Baptist University \\
$^4$Data Science Research Center, Duke Kunshan University \\
$^5$Shanxi Provincial People's Hospital \\
$^6$The Fifth Clinical Medical school of Shanxi Medical University\\
$^7$Research Center for Mathematics, Beijing Normal University \\
$^8$Department of Mathematics and Statistics, York University \\
\\
}
\date{} 
\begin{document}
\maketitle
\begin{abstract}
Accurate medical image analysis can greatly assist clinical diagnosis, but its effectiveness relies on high-quality expert annotations. Obtaining pixel-level labels for medical images, particularly fundus images, remains costly and time-consuming. Meanwhile, despite the success of deep learning in medical imaging, the lack of interpretability limits its clinical adoption. To address these challenges, we propose TWLR, a two-stage framework for interpretable diabetic retinopathy (DR) assessment. In the first stage, a vision–language model integrates domain-specific ophthalmological knowledge into text embeddings to jointly perform DR grading and lesion classification, effectively linking semantic medical concepts with visual features. 
The second stage introduces an iterative severity regression framework based on weakly-supervised semantic segmentation. Lesion saliency maps generated through iterative refinement direct a progressive inpainting mechanism that systematically eliminates pathological features, effectively downgrading disease severity toward healthier fundus appearances. Critically, this severity regression approach achieves dual benefits: accurate lesion localization without pixel-level supervision and providing an interpretable visualization of disease-to-healthy transformations. Experimental results on the FGADR, DDR, and a private dataset demonstrate that TWLR achieves competitive performance in both DR classification and lesion segmentation, offering a more explainable and annotation-efficient solution for automated retinal image analysis.
\end{abstract}


\section{Introduction}

Diabetic Retinopathy (DR), a common complication of diabetes, serves as the leading cause of blindness worldwide. Progressive visual loss is primarily caused by long-term high blood sugar levels, which damage the retinal microvasculature, resulting in vascular leakage or obstruction and a range of retinal lesions \cite{fong20043rd}. In clinical practice, ophthalmologists typically assess the symptoms of the retina through fundus images to evaluate the progression and prognosis of the disease. Recent advancements in deep learning have facilitated the development of automated diagnostic systems, aiding ophthalmologists in disease grading and lesion localization on fundus images, with convolutional neural networks (CNNs) and Transformer-based models serving as key architectures \cite{he2020cabnet,hanselmann2020canet,huang2021lesion,yu2021mil,wen2024concept}. However, clinical adoption faces two primary challenges. First, a vast amount of fundus images remain underutilized, as most are annotated only at the image level with severity grades, lacking the pixel-level lesion annotations that are costly and time-intensive to produce. Second, existing deep learning models for DR grading act as ``black boxes," providing only severity predictions without elucidating whether lesions drive the diagnosis. This lack of interpretability hinders ophthalmologists from validating model decisions and limits clinical trust and adoption. To address both the annotation cost and interpretability challenges, recent weakly-supervised methods \cite{gondal2017weakly,playout2019novel,zhao2021robust,gonzalez2020iterative,peng2024hierarchical} leverage Class Activation Maps (CAMs) \cite{b10}, Grad-CAM \cite{b11} to localize lesions and provide visual explanations using only image-level labels. By generating saliency maps that highlight diagnostically relevant regions, these approaches offer a degree of interpretability while reducing annotation burden. However, these approaches face critical limitations in fundus image analysis. First, traditional weakly-supervised semantic segmentation (WSSS) techniques are designed to capture large, salient regions. This makes them poorly suited for retinal lesions that exhibit diverse presentations, ranging from tiny microaneurysms to large, irregular hemorrhages and exudates. Small lesions with blurred boundaries and low contrast are particularly challenging to detect. Consequently, existing methods tend to highlight dominant features while missing smaller, yet clinically significant lesions, limiting their ability to provide comprehensive clinical visual explanations. Second, the robustness and validity of these methods' reasoning remain underexplored. While they generate saliency maps, it is unclear whether the models genuinely rely on lesion features for grading or exploit spurious image 
characteristics. The effect of targeted perturbations—including pixel-level changes and lesion-level manipulations on DR grading accuracy and localization performance has not been systematically investigated. Such analysis is essential to verify that model predictions are driven by clinically meaningful lesion features.

To address these challenges in WSSS and model interpretability, we propose $\textbf{T}$ext-guided $\textbf{W}$eakly-supervised $\textbf{L}$esion localization and severity $\textbf{R}$egression ($\textbf{TWLR}$), a two-stage text-guided vision-language framework that integrates multimodal DR grading and lesion classification with weakly-supervised iterative lesion localization for explainable diabetic retinopathy diagnosis. In the first stage, we develop a text-guided vision-language framework based on Contrastive Language-Image Pre-Training (CLIP) architecture \cite{b18}. By explicitly aligning textual clinical descriptions—including specialized medical terminology for lesion types and DR severity grading criteria with fundus images, our approach enables the model to understand domain-specific ophthalmological concepts and leverage expert knowledge to guide both DR grading and lesion classification. In the second stage, to achieve comprehensive lesion localization, we propose an iterative refinement mechanism that progressively identifies all lesion regions through a closed-loop process. After obtaining the initial DR grading from the vision-language model (Stage 1), we employ guided backpropagation to generate saliency maps that highlight lesion-related regions contributing to the classification decision. These identified lesion areas are then localized and subsequently ``restored" using the LaMa inpainting method \cite{b19}, simulating a disease regression process where the fundus image transitions toward a healthier state with reduced pathological severity. The restored image is then fed back into the multimodal DR grading model for reassessment, forming an iterative loop: DR grading → lesion localization → inpainting-based restoration → regrading. This loop continues until the model classifies the restored fundus image as healthy, ensuring that all clinically significant lesion regions have been comprehensively identified and localized throughout the iterative process. Through this iterative text-guided framework, TWLR not only achieves accurate weakly-supervised lesion localization but also provides multi-level explainability for DR diagnosis. By combining vision-language alignment with progressive lesion identification, our method generates both textual explanations (identifying specific lesion types present) and visual explanations (spatially localizing lesions), offering ophthalmologists comprehensive, clinically interpretable diagnostic insights.

Our contributions are summarised as follows:

$\bullet$ We introduce TWLR, an iterative text-guided weakly-supervised localization strategy that leverages vision-language alignment to comprehensively identify all lesion regions by mimicking the disease regression process. By integrating textual lesion descriptions with visual features, our method employs text prompts to guide the model's attention toward specific lesion types at each severity level. Through repeated grading-localization-repair cycles, TWLR progressively detects lesions guided by their corresponding textual semantics, ensuring comprehensive identification of all DR lesion regions without pixel-level annotations.

$\bullet$ Extensive experiments on DDR, FGADR, and our private datasets demonstrate that TWLR achieves competitive performance in DR grading and lesion classification while providing clinically meaningful visual and textual explanations, validating its effectiveness for explainable weakly-supervised DR diagnosis.



\section{Related Work}
\subsection{Deep Learning for Diabetic Retinopathy Classification}
Deep learning has transformed medical imaging analysis due to its exceptional ability to automatically learn complex visual patterns from large datasets without explicit feature engineering. In DR grading, these techniques have especially enabled significant advances in the automated detection and classification of disease severity from retinal fundus photographs.
In their pivotal work, early studies \cite{b20, b21} developed and validated on Convolution Neural Networks (CNN)-based architecture such as Inception-V3, VGG16 to detect DR. Building on this, He et al. \cite{b22} introduced CABNet, which incorporates a Category Attention Block within a CNN framework to enhance the model's sensitivity to minority classes.  

Recent advancements in DR grading have shifted from pure CNN-based approaches to methods that explicitly integrate lesion-specific information and saliency guidance. Building on traditional CNN architectures, researchers now focus on encoding lesion-aware features to improve both accuracy and interpretability. Hou et al. \cite{b23} pioneered the integration of weakly supervised lesion priors into DR grading systems, using CAM-based methods to highlight clinically relevant regions without requiring expensive pixel-level annotations. This approach significantly reduced the annotation burden while maintaining diagnostic accuracy comparable to that of fully supervised methods. Complementing this work, Huang et al. \cite{huang2021lesion} introduced lesion-based contrastive learning, which explicitly contrasts lesion-positive and lesion-negative image pairs during training. For multi-task retinal diagnosis, Li et al. \cite{b25} designed CANet, a cross-disease attention network that leverages complementary information from both DR and diabetic macular edema. By implementing a dual-branch architecture with shared lesion-specific features and disease-specific attention modules, CANet captures the internal relationships between these two diseases.

Unlike previous approaches, our model leverages the Vision Transformer (ViT) architecture \cite{wu2020visual}, which establishes relationships between arbitrary positions within the image through its self-attention mechanism. This enables global feature capture that transcends the limited receptive fields of traditional CNNs. Additionally, we incorporate textual expert fundus knowledge by injecting embeddings of DR severity levels and their corresponding lesion descriptors, aligning this domain expertise with visual features. This novel integration enables the model to comprehend and identify both the severity grades and specific lesion patterns in fundus images. Compared to conventional one-hot encoding approaches for DR classification, our method maintains strong generalization capabilities even when classification categories change, as it learns meaningful representations of the underlying lesion concepts rather than merely mapping to predefined class labels. This knowledge-guided approach bridges the semantic gap between visual patterns and clinical interpretations, resulting in more robust and clinically relevant feature extraction.
\subsection{Weakly Supervised Semantic Segmentation in Medical Imaging}
Due to the limited availability of expert-annotated medical image data, WSSS has emerged as a crucial and rapidly evolving direction in deep learning. Unlike traditional approaches that require pixel-level annotations, WSSS methods can leverage weaker forms of supervision such as image-level labels, bounding boxes, or scribbles to achieve segmentation tasks \cite{chen2022a, chen2022b, ma2024segment, zhong2024weakly, wang2025weakmedsam}. In the field of medical imaging, this paradigm is particularly valuable as it significantly reduces annotation burden while still enabling precise localization of medical abnormalities. Particularly, weakly supervised learning has also been investigated in the lesion localization of fundus images. Gonzalo et al. \cite{gonzalez2020iterative} proposed an iterative visual evidence augmentation approach that progressively refines lesion localization by integrating interpretability frameworks with weakly supervised learning. Gondal et al. \cite{gondal2017weakly} developed a framework for weakly supervised localization of DR lesions in retinal fundus images by leveraging attention mechanisms to identify discriminative regions without pixel-level annotations. Morano et al.\cite{b14} introduced an explainable deep learning approach for weakly supervised detection of age-related macular degeneration lesions in color fundus images, enabling both lesion classification and model decision transparency. Similarly, Li et al. \cite{b28} presented a method for WSSS of DR lesions that effectively translates image-level diagnostic labels into detailed lesion segmentation maps.\\
\hspace*{1em}Despite these advances, weakly supervised lesion segmentation in fundus images remains particularly challenging. The main challenges come from unclear lesion boundaries, highly variable lesion shapes, and their scattered patterns across the fundus. These factors significantly limit the effectiveness of weak supervision, calling for more sophisticated techniques tailored to medical image analysis.\\
\hspace*{1em}Building upon the framework introduced by Gonzalo et al. \cite{b26}, we explore and extend their iterative approach by incorporating advanced vision-language models and improved inpainting techniques for DR lesion localization. This method systematically identifies key lesions from DR classification, inpaints the detected lesions, and then feeds the inpainted images with reduced severity back to the classification model for subsequent iterations. To implement this classification-localization-inpainting iterative process, we leverage a vision-language model derived from CLIP \cite{b18} for weakly supervised lesion segmentation, enabling more precise localization of lesion regions without pixel-level annotations. For the inpainting phase, we employ LaMa \cite{b19}, a state-of-the-art inpainting model, to replace identified lesion regions with realistic healthy fundus textures, thereby enabling the progressive removal of disease. This iterative mechanism delivers two key advantages: first, it enables comprehensive lesion detection by progressively removing prominent abnormalities to expose subtler, previously hidden lesions; second, it demonstrates the transformation from diseased to healthy fundus states while preserving critical anatomical structures, particularly retinal vasculature. Through this process, our approach not only enhances diagnostic accuracy but also provides interpretable visualizations of disease progression.

\begin{figure*}[htbp]
    \centering
    \includegraphics[width=1.0\linewidth]{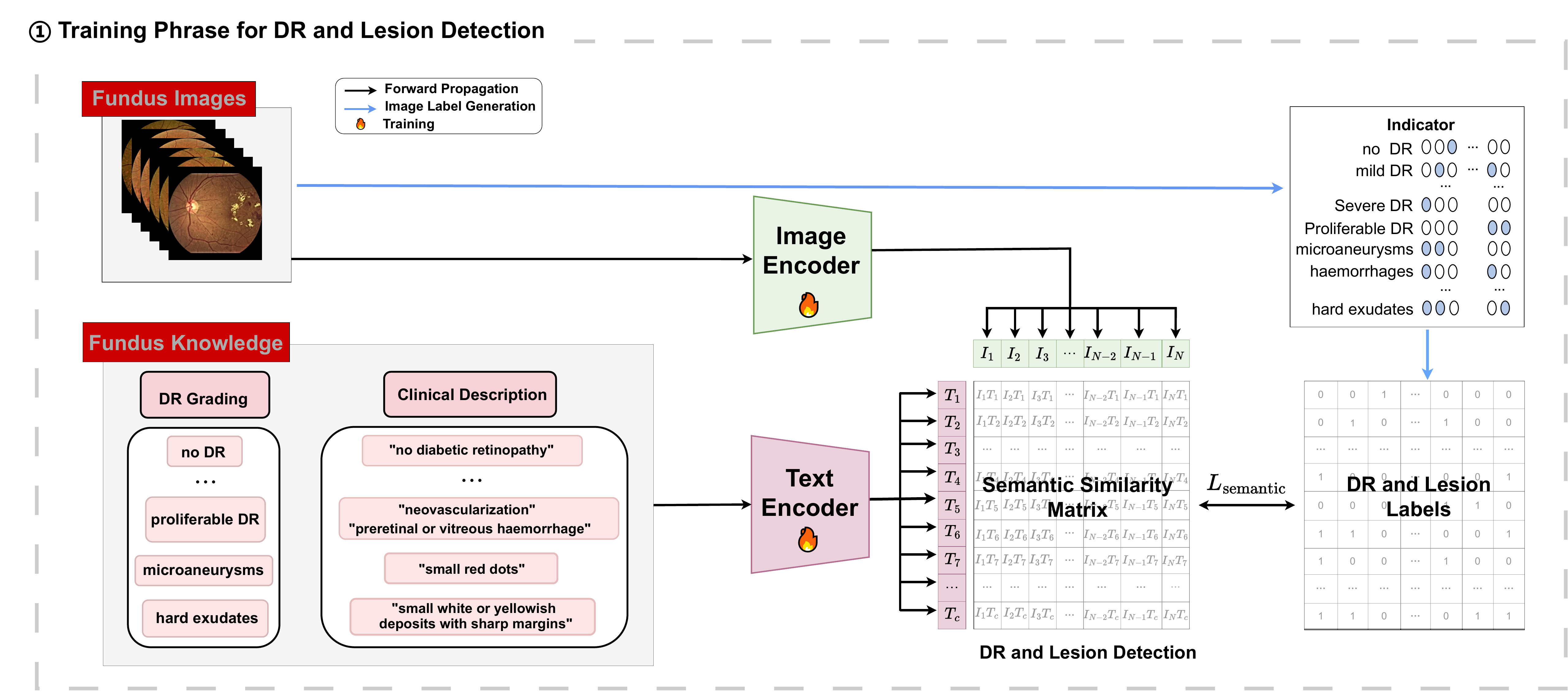}
     \caption{Training phase of the TWLR model for multi-task DR grading and lesion classification. The vision-language multi-modal framework processes fundus images through an image encoder to extract visual embeddings. Instead of traditional categorical labels, we employ detailed lesion descriptions (e.g., "only few microaneurysms", "neovascularization", "small red dots") as textual inputs that replace conventional DR grading categories and lesion labels. These descriptions are encoded and interact with visual features to generate a semantic similarity matrix, which serves as the final prediction. Training labels are constructed with a hybrid structure combining DR severity levels and binary lesion presence indicators, enabling effective multi-task learning through description-based visual-textual alignment.}
    \label{fig:iterative_DR_training}
\end{figure*}

\section{Methodology}
\begin{figure*}[!tbp]
    \centering
    \includegraphics[width=1.0\linewidth]{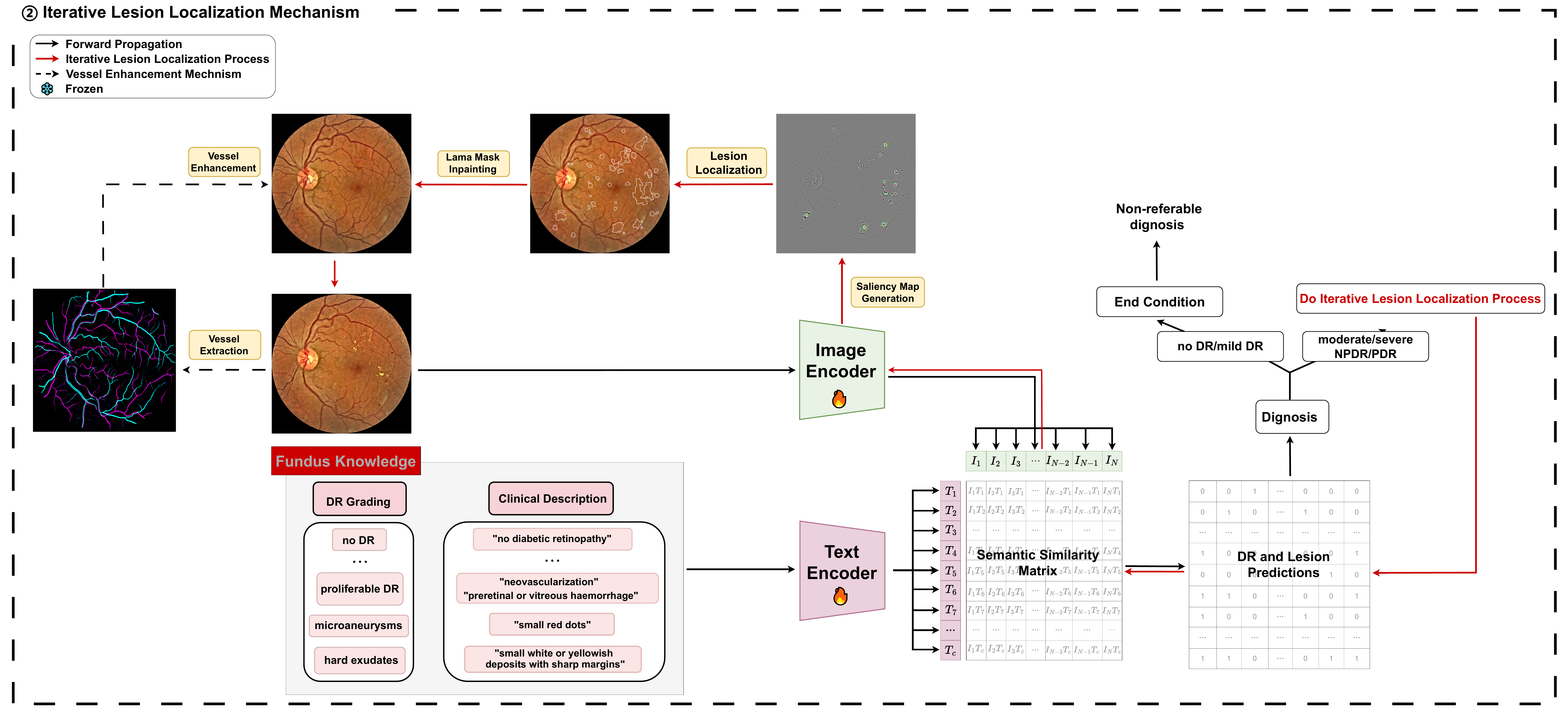}
    \caption{Testing phase of the TWLR model for severity regression-based lesion localization process. Fundus images are first fed into the multi-modal framework to obtain DR grading and lesion predictions. If a fundus image is classified as ``no DR'', it receives a non-referable diagnosis. Otherwise, images detected with DR are introduced to the severity regression-based lesion localization process. This process employs multiple steps: (1) Initial multi-modal classification generates predictions and saliency maps via guided backpropagation to identify diagnostically relevant regions; (2) Thresholding converts saliency maps to binary representations highlighting salient areas; (3) LaMa inpainting technique reconstructs and refines these regions, utilizing fast Fourier convolutions for high-quality restoration while preserving anatomical details. This iterative cycle enables weakly supervised lesion segmentation using only image-level labels, progressively improving diagnostic accuracy and model interpretability through continuous refinement of attention mechanisms focused on clinically significant retinal structures.}
    \label{fig:iterative_DR_testing}
\end{figure*}

Fig. \ref{fig:iterative_DR_training} and Fig. \ref{fig:iterative_DR_testing} illustrate the comprehensive architecture of our proposed Text-guided Weakly-supervised Lesion localization and severity Regression (TWLR) framework. Fig. \ref{fig:iterative_DR_training} shows the training phase of DR and lesion classification based on a vision-language model, and Fig. \ref{fig:iterative_DR_testing} shows the test phase of weakly supervised lesion segmentation and inpainting based on the severity regression-based lesion localization process. The system consists of two principal components: (1) Fig. \ref{fig:iterative_DR_training} shows that our vision-language DR detection module integrates specialized ophthalmological knowledge about DR grading standards and lesion descriptions with fundus image analysis. This integration enables robust and clinically relevant DR detection through multimodal alignment between textual expertise and visual features. (2) Fig. \ref{fig:iterative_DR_testing} illustrates our comprehensive lesion localization framework that enhances both detection accuracy and clinical interpretability. The process begins with vision-language model-based DR grading: for cases classified as No DR or Mild DR, the system directly outputs a non-referable diagnosis, requiring no further analysis. However, for moderate to severe DR cases that require clinical referral, the framework initiates an iterative refinement process. This iterative pipeline operates as follows: after identifying referable DR, we extract saliency maps highlighting potential lesion regions through weakly supervised segmentation. These regions undergo precise localization followed by advanced inpainting to computationally restore affected areas to healthy appearances. The reconstructed image is then re-evaluated by the vision-language model, creating a feedback loop that progressively uncovers additional lesions. 

\subsection{Vision-language Model for DR Diagnosis}\label{sectionA}
The integration of domain-specific medical knowledge into vision-based diagnostic systems represents a significant advancement in computer-aided medical image analysis. Injecting expert ophthalmological knowledge into vision models has emerged as a particularly effective approach for enhancing representational capabilities and diagnostic accuracy in clinical fundus decisions. Unlike conventional vision-only architectures that rely solely on image features, multimodal frameworks establish meaningful correlations between standardized medical concepts and corresponding lesion manifestations within medical images, facilitating more robust and clinically relevant diagnostic outcomes. \\
\hspace*{1em} Inspired by a retina vision-language foundation model proposed in \cite{b18}, our approach leverages comprehensive expert knowledge descriptors specifically tailored to DR grading standards and lesion description. This multimodal framework creates a robust alignment between fundus images and specialized clinical textual descriptions, enabling more accurate and interpretable DR assessments.

Let $\mathcal{D} = \{(I^{(i)}, y^{(i)})\}_{i=1}^{N_D}$ denote a fundus image dataset with $N_D$ samples, where for each sample, $I^{(i)} \in \mathbb{R}^{H \times W \times C}$ represents the fundus image with spatial dimensions $H \times W$ and $C$ channels, and $y^{(i)} =[y_{DR}^{(i)}, y_{lesion}^{(i)}]\in \mathbb{R}^{\mathcal{C}_{grad}+\mathcal{C}_{lesion}}$ denotes the composite DR grading and lesion labels for $i$-th fundus image. Specifically, the DR grading label for $i$-th fundus image $y_{DR}^{(i)} \in \mathbb{R}^{\mathcal{C}_{grad}}$ is a one-hot vector encoding the severity level from $\mathcal{C}_{grad}$ = \{``No DR'', \text{``Mild DR''}, ``Moderate DR'', ``Severe DR'', ``Proliferative DR''\}, and $y_{lesion}^{(i)} \in \mathbb{R}^{\mathcal{C}_{lesion}}$ is a multi-label binary vector for the $i$-th indicating the presence of lesion types from set $\mathcal{C}_{lesion}$ = \{``MA'', ``HE'', ``SE'', ``EX''\}, representing microaneurysms, hemorrhages, soft exudates, and hard exudates respectively. Corresponding to these label sets, we define clinical descriptions: $\{T_{j}^{grad}\}_{j=1}^{|C_{grad}|}$ for DR grades and $\{T_{k}^{lesion}\}_{k=1}^{|C_{lesion}|}$ for lesion types. For example, ``Mild DR'' in $\mathcal{C}_{grad}$ is characterized by ``only few microaneurysms'', while lesion type ``MA'' in $\mathcal{C}_{lesion}$ is described as ``small red dots''. These descriptions are derived from international DR clinical standards \cite{wilkinson2003proposed} and established clinical literature \cite{garner1979pathogenesis}. Note that each level in $\mathcal{C}_{grad}$ or $\mathcal{C}_{lesion}$ may be associated with multiple text descriptions to capture diverse clinical manifestations. The objective is to train a multi-modal framework to obtain DR and lesion prediction: $f_{\text{VL}}: I^{(i)} \rightarrow \hat{y}^{(i)}$.

As illustrated in Fig.~\ref{fig:iterative_DR_training}, our approach fundamentally differs from traditional multi-class classification methods. Instead of relying on conventional multi-hot binary labels for DR grading and lesion detection, we leverage rich textual embeddings derived from clinical descriptions as supervision signals. This design enables the model to learn nuanced relationships between visual features and medical semantics.

Specifically, we replace discrete category labels with comprehensive lesion descriptions $\{T_{j}^{grad}\}_{j=1}^{|\mathcal{C}_{grad}|}$ and $\{T_{k}^{lesion}\}_{k=1}^{|\mathcal{C}_{lesion}|}$ that capture clinical characteristics of each DR grade and lesion type. Each description is processed through the text encoder to obtain corresponding textual representations:
\begin{align}
& \mathbf{E}_{text}^{grad}  = \{f_{text}(T_{j}^{grad}) \in \mathbb{R}^D \mid j=1,...,|\mathcal{C}_{grad}|\} \\
& \mathbf{E} _{text}^{lesion} = \{f_{text}(T_{k}^{lesion}) \in \mathbb{R}^D \mid k=1,...,|\mathcal{C}_{lesion}|\}\\
& \mathbf{E}_{text} = [\mathbf{E}_{text}^{grad}, \mathbf{E}_{text}^{lesion}] \in \mathbb{R}^{(\mathcal{C}_{grad}+\mathcal{C}_{lesion}) \times D}
\end{align}

where $f_{text}(\cdot)$ represents the text encoder that maps each clinical description to a $D$-dimensional embedding space. $ E_{text}^{grad}$ and $E_{text}^{lesion}$ denote all the embeddings of the textual description of DR gradings and lesions. Rather than using abstract class labels alone, we leverage detailed textual descriptions of DR severity grades and lesions to perform downstream tasks of fundus image classification. This approach allows the model to develop a richer understanding of clinical terminology by associating visual patterns with their corresponding medical descriptions. For instance, instead of simply learning that a region contains ``microaneurysms'', the model learns that they are manifested as ``small red dots'' bridging the gap between clinical terminology and visual appearance. 

For a batch of fundus images $\{I^{(i)}\}_{i=1}^{N_{batch}}$, we employ the ViT architecture \cite{dosovitskiy2020image} as the image encoder to extract discriminative visual features for DR diagnosis. Given an input fundus image $I^{(i)} \in \mathbb{R}^{H \times W \times C}$, we first partition it into patches:
\begin{align}
    \text{split}(I^{(i)}) = \{\mathbf{p}_j\}_{j=1}^N, \; \mathbf{p}_j \in \mathbb{R}^{P\times P \times C},\;  N = HW/P^{2}
    \label{eq4}
\end{align}

The patch embedding process is then formulated as:
\begin{align}
\mathbf{x}_j &= \text{Flatten}(\mathbf{p}_j) \in \mathbb{R}^{1\times P^2C}, \quad j = 1, \ldots, N \\
\mathbf{z}_j^{(0)} &= \mathbf{x}_j \mathbf{W} + \mathbf{e}_{pos}^{(j)} \in \mathbb{R}^{1\times D}
\end{align}
where $\mathbf{p}_j \in \mathbb{R}^{P \times P \times C}$ represents the $j$-th image patch. $\mathbf{x}_j \in \mathbb{R}^{1 \times P^2C}$ is the flattened row vector of $\mathbf{p}_j$. $\mathbf{W} \in \mathbb{R}^{(P^2C) \times D}$ is the learnable linear projection matrix that maps the $P^2C$-dimensional flattened patches $\mathbf{x}_j$ to $D$-dimensional embeddings. $\mathbf{e}_{pos}^{(j)} \in \mathbb{R}^{1 \times D}$ denotes the positional embedding for the $j$-th patch.

A learnable classification token $\mathbf{z}_{cls}^{(0)} \in \mathbb{R}^{1\times D}$ is prepended to the patch sequence, forming the complete input sequence:

\begin{equation}
\mathbf{Z}^{(0)} = [\mathbf{z}_{cls}^{(0)}, \mathbf{z}_1^{(0)}, \mathbf{z}_2^{(0)}, \ldots, \mathbf{z}_N^{(0)}] \in \mathbb{R}^{(N+1) \times D}
\end{equation}

This sequence $\mathbf{Z}^{(0)}$ are fed into a Transformer encoder consisting of \(L\) layers. The process through these layers can be summarized as:
\begin{align}
& \mathbf{Z}^{(L)} = f_{\text{Transformer}}(\mathbf{Z}^{(0)}) \in \mathbb{R}^{(N+1)\times D}\\
& \mathbf{E}_{img}^{(i)} =\mathbf{z}_{cls}^{(L)} \in \mathbb{R}^{1\times D}\label{eq9}
\end{align}

where \(f_{\text{Transformer}}(\cdot)\) represents the combined operations of self-attention and feed-forward networks applied iteratively across \(L\) layers.

After processing, the final output $\mathbf{Z}^{(L)}\in \mathbb{R}^{(N+1)\times D}$ includes the embeddings for all patches, from which we can extract a special classification token $\mathbf{z}_{cls}^{(L)}\in \mathbb{R}^{1\times D}$ as the global representation for the image $\mathbf{E}_{img}^{(i)}$.

Each image from a batch of fundus images $\{I^{(i)}\}_{i=1}^{N_{batch}}$ is processed 
through Eq. \ref{eq4}-\ref{eq9} to obtain its corresponding image representation. To predict the DR grading and lesion types for each image, we compute the similarity between text and image embeddings across all semantic classes. After obtaining the text and image embeddings, we calculate the similarity between two modalities as the predictions of the DR grading and lesion types for the target images.
Specifically, 
$\mathbf{E}_{text} \in \mathbb{R}^{(\mathcal{C}_{grad}+\mathcal{C}_{lesion}) \times D}$ represents the 
concatenated textual embeddings for DR grading levels and lesion descriptions, where each row corresponds to a semantic class. For the $i$-th fundus image with embedding $\mathbf{E}_{img}^{(i)} \in \mathbb{R}^{D}$, we compute the prediction score $\hat{y}^{(i)}$ and and minimize the semantic loss $\mathcal{L}_{\text{semantic}}$:
\begin{align}
&\hat{\mathbf{s}}^{(i)} = \mathbf{E}_{text}\cdot\mathbf{E}_{img}^{(i)\top}\in \mathbb{R}^{C_{grad}+C_{lesion}}\\
&\hat{y}^{(i)} = \frac{exp(\tau\hat{ \mathbf{s}}^{(i)})}{\sum_{i=1}^{N_{batch}}exp(\tau \hat{\mathbf{s}}^{(i)} )}\\
& {y}^{(i)} = \frac{y^{(i)}}{\|y^{(i)}\|_1} 
\end{align}

\begin{equation}
\mathcal{L}_{\text{semantic}} = -\frac{1}{N_{batch}}\sum_{i=1}^{N_{batch}} y^{(i)} \odot\log(\hat{y}^{(i)} ).
\end{equation}

where $\hat{\mathbf{s}}^{(i)} \in \mathbb{R}^{\mathcal{C}_{grad}+\mathcal{C}_{lesion}}$ is the 
similarity score vector, $\hat{y}^{(i)}$ is the predicted score of each class, $\tau$ is the learnable temperature parameter, $y^{(i)}$ is the multi-hot ground truth vector, indicating indicates the presence or absence of each DR grading and lesion class. Normalization is to ensure consistency and comparability during training. Finally, we use the $\mathcal{L}_{\text{semantic}}$ to drive model training for DR grading and lesion classification.


\begin{algorithm}[t]
\caption{Severity Regression-based Lesion Localization Process}
\label{alg:iterative_lesion_loc}
\begin{algorithmic}[1]
\REQUIRE Batch of fundus images $\{I^{(i)}\}_{i=1}^{N_{\text{batch}}}$, DR prediction model $f_{\text{VL}}(\cdot; \theta)$, inpainting model $\text{LaMa}(\cdot; \phi)$, threshold $\tau$
\ENSURE Final inpainted images $\{I^{(i,T_i)}\}_{i=1}^{N_{\text{batch}}}$, lesion masks $\{I_{\text{lesion}}^{(i,T_i)}\}_{i=1}^{N_{\text{batch}}}$
\STATE Define $\mathcal{C}_{\text{non-referable}} = \{0, 1\}$ (No DR and Mild DR)
\STATE Define $\mathcal{C}_{\text{referable}} = \{2, 3, 4\}$ (Moderate, Severe, Proliferative DR)
\FOR{$i = 1$ to $N_{\text{batch}}$}
    \STATE $t \leftarrow 1$
    \STATE $I^{(i,t)} \leftarrow I^{(i)}$ \hfill // Initialize with original image
    \STATE $I_{\text{lesion}}^{(i,t-1)} \leftarrow \emptyset$ \hfill // Initialize empty lesion mask
    \REPEAT
        \STATE $\hat{y}^{(i,t)} \leftarrow f_{\text{VL}}(I^{(i,t)}; \theta)$ \hfill // Get DR grade probabilities
        \STATE $\hat{c}^{(i,t)} \leftarrow \arg\max_{c} \hat{y}^{(i,t)}$ \hfill // Get predicted class
        \IF{$\hat{c}^{(i,t)} \in \mathcal{C}_{\text{referable}}$}
            \STATE $S^{(i,t)} \leftarrow \text{GuidedBackprop}(I^{(i,t)}, \hat{y}^{(i,t)}; \theta)$ \hfill // Generate saliency map
            \STATE $I_{\text{binary}}^{(i,t)} \leftarrow \mathbb{I}_{S^{(i,t)} > \tau}$ \hfill // Binarize saliency map
            \STATE $I_{\text{lesion}}^{(i,t)} \leftarrow I_{\text{lesion}}^{(i,t-1)} \cup I_{\text{binary}}^{(i,t)}$ \hfill // Accumulate lesions
            \STATE $I^{(i,t+1)} \leftarrow \text{LaMa}(I^{(i,t)}, I_{\text{binary}}^{(i,t)}; \phi)$ \hfill // Inpaint lesions
            \STATE $t \leftarrow t + 1$
        \ENDIF
    \UNTIL{$\hat{c}^{(i,t)} \in \mathcal{C}_{\text{non-referable}}$}
    \STATE $T_i \leftarrow t$ \hfill // Record total iterations for image $i$
\ENDFOR
\RETURN $\{I^{(i,T_i)}\}_{i=1}^{N_{\text{batch}}}, \{I_{\text{lesion}}^{(i,T_i)}\}_{i=1}^{N_{\text{batch}}}$
\end{algorithmic}
\end{algorithm}




\subsection{Severity Regression-based Lesion Localization Process}

In addition to the primary task of vision-language multi-modal DR grading and lesion classification, we investigate lesion localization under a WSSS framework using only image-level labels. This approach aims to identify the spatial locations of DR-related lesions without requiring labor-intensive pixel-level annotations, thereby substantially reducing the annotation burden while maintaining clinically valuable localization performance. Moreover, localizing lesions that contribute to the diagnostic decision enhances the clinical interpretability of our model by explicitly revealing which pathological features drive the predictions.

To achieve this, we propose a severity regression-based lesion localization strategy that leverages the hierarchical relationship between DR severity grades and their underlying lesion patterns. Specifically, given a fundus image and its predicted severity grade, our model traces back from the grade prediction to identify the discriminative lesion regions that most strongly support the diagnosis. This is accomplished through attention mechanisms that capture the most relevant regions during the grading process, which are subsequently refined to generate pixel-level lesion localization maps. 

Moreover, once a set of lesions is localized, we perform inpainting to remove these identified lesions from the image and re-grade the restored fundus image. This iterative grading-localization-restoration cycle enables our model to progressively discover all lesions at different severity levels: after removing the most prominent lesions, the model can focus on detecting the remaining, potentially less obvious lesions that were previously overshadowed. This iterative process continues until all lesion regions are comprehensively identified. Unlike conventional fully supervised segmentation methods, our weakly supervised approach exploits the inherent correlation between image-level diagnostic labels and lesion manifestations, eliminating the need for dense pixel-wise annotations.

This mechanism integrates predictions with visual explanations to accomplish comprehensive lesion localization through severity regression, as illustrated in Algorithm \ref{alg:iterative_lesion_loc} and Fig. \ref{fig:iterative_DR_testing}. We categorize DR severity into two groups: 
non-referable grades ($\mathcal{C}_{\text{non-referable}} = \{0, 1\}$, including 
No DR and Mild DR) and referable grades ($\mathcal{C}_{\text{referable}} = \{2, 3, 4\}$, 
comprising Moderate, Severe, and Proliferative DR). For each fundus image $I^{(i)}$ in the batch, we initialize the iteration counter 
$t=1$, set the current image as $I^{(i,1)} \leftarrow I^{(i)}$, and initialize the lesion mask as empty, $I_{\text{lesion}}^{(i,0)} \leftarrow \emptyset$.
The algorithm iteratively processes the image through our vision-language model 
$f_{\text{VL}}$ to obtain DR grade probabilities $\hat{y}^{(i,t)}$ and the predicted DR severity grade $\hat{c}^{(i,t)}$.

\begin{algorithm}[t]
\caption{Vessel Enhancement and Repair Process}
\label{alg:vessel_repair}
\begin{algorithmic}[1]
\REQUIRE Inpainted image $I^{(i,t)}$, original image $I^{(i,1)}$, vessel segmentation $V_{\text{seg}}^{(i)}$, identified lesion mask $I_{\text{lesion}}^{(i,t-1)}$, vessel threshold $\tau_v$, blending weight for vessel $\alpha_{\text{vessel}}$, blending weight for the intersection between identified lesion regions and vessels $\alpha_{\text{inter}}$
\ENSURE Vessel-enhanced image $\{I_{\text{enhance}}^{(i,t)}\}_{i=1}^{N_{batch}}$
\FOR{$i = 1$ to $N_{\text{batch}}$}
    \STATE \textbf{Phase 1: Vessel Mask Preprocessing}
    \STATE $V_{\text{seg}}^{(i)}=f_{\text{seg}}(I^{(i,1)})$ // Get predicted vessel mask
    \STATE $V_{\text{prerpocess}}^{(i)} \leftarrow \text{Preprocess}(V_{\text{seg}}^{(i)})$ // Preprocess predicted vessel mask
    
    \STATE
    \STATE \textbf{Phase 2: Intersection Between Identified Regions and Vessels}
    \STATE $V_{\text{inter}}^{(i,t)} \leftarrow I_{\text{binary}}^{(i,t)} \cap V_{\text{preprocess}}^{(i)}$ \hfill // Find intersections with inpainted region
    
    \STATE
    \STATE \textbf{Phase 3: Color Matching and Blending}
    \STATE $\mu_c^{(i)}, \sigma_c^{(i)} \leftarrow \text{ExtractVesselColor}(I^{(i,1)}, V_{\text{preprocess}}^{(i)})$ \hfill // Extract mean and standard error from original image
    \STATE $V_{\text{color}}^{(i)} \leftarrow \text{GenerateColoredVessels}(V_{\text{preprocess}}^{(i)}, \mu_c^{(i)}, \sigma_c^{(i)})$ // See details in Algorithm 3
    
    \STATE
    \STATE \textbf{Phase 4: Apply blending with enhanced weight in intersection regions}
    
    \FOR{each pixel location $(x,y)$}
        \IF{$V_{\text{inter}}^{(i,t)}(x,y) > 0$}
            \STATE $I_{\text{enhance}}^{(i,t)}(x,y) \leftarrow (1-\alpha_{\text{inter}}) I^{(i,t)}(x,y) + \alpha_{\text{inter}} V_{\text{color}}^{(i)}(x,y)$ 
        \ELSE
            \STATE $I_{\text{enhance}}^{(i,t)}(x,y) \leftarrow (1-\alpha_{\text{vessel}}) I^{(i,t)}(x,y) + \alpha_{\text{vessel}} V_{\text{color}}^{(i)}(x,y)$ 
        \ENDIF
    \ENDFOR
\ENDFOR    
\STATE
\RETURN $\{I_{\text{enhance}}^{(i,t)}\}_{i=1}^{N_{batch}}$
\end{algorithmic}
\end{algorithm}

When a referable grade is predicted, we employ Guided Backpropagation \cite{b13} to generate 
a saliency map $S^{(i,t)}$ that reveals the regions most influential to the model's classification decision. To obtain a binary representation of lesion areas, we apply a thresholding technique with threshold $\tau$ that converts the continuous saliency 
values into binary masks $I_{\text{binary}}^{(i,t)} = \mathbb{I}_{S^{(i,t)} > \tau}$, where $\tau$ is defined as the mean plus one standard deviation of the pixel values in saliency map $S^{(i,t)}$. This binary map highlights the most salient regions, which typically correspond to the most discriminative features that the model relies on for the predicted DR class. Specifically, these highlighted areas represent lesion patterns that are most characteristic of the predicted DR severity level, such as microaneurysms and hemorrhages for moderate DR, or neovascularization for proliferative DR. By identifying these model-critical regions, we effectively localize lesion areas 
that are most responsible for the DR classification decision. We accumulate all identified lesion regions across iterations as $I_{\text{lesion}}^{(i,t)} = I_{\text{lesion}}^{(i,t-1)} \cup I_{\text{binary}}^{(i,t)}$.

\begin{algorithm}[!t]
\caption{Generate Colored Vessels with Relative Distance}
\label{alg:generate_colored_vessels}
\begin{algorithmic}[1]
\REQUIRE Vessel mask $V_{\text{preprocess}}^{(i)}=\{v \in \mathbb{R}^{H \times W} \mid v_{ij} \in \{0,1\}\}$, color mean $\mu_c^{(i)} \in \mathbb{R}^3$, color std $\sigma_c^{(i)} \in \mathbb{R}^3$, darkening factor $\beta_{\text{dark}}$, relative distance factor $\gamma_{\text{distance}}$, noise scaling $\delta_{\text{noise}}$

\ENSURE Colored vessel image  \ensuremath{V_{\text{color}}^{(i)} = \{m\in \mathbb{R}^{H \times W \times 3}|m_{ij} \in [0,255]\}}

\STATE \textbf{Step 1: Initialization}
\STATE $V_{\text{color}}^{(i)} \leftarrow \mathbf{O}_{H \times W \times 3}$ \hfill // Initialize output
\STATE $\mathcal{V} \leftarrow \{(x,y) \mid v_{xy} = 1, \text{ where } v_{xy} \triangleq V^{(i)}_{\text{preprocess}}[x, y] \}$ \hfill // Extract vessel coordinates

\STATE
\STATE \textbf{Step 2: Compute Base Color}
\STATE $C^{(i)}_{\text{base}} \leftarrow \beta_{\text{dark}} \cdot \mu_c^{(i)}$ \hfill // Darken base color

\STATE
\STATE \textbf{Step 3: Compute Geometric Center}
\STATE $(c_x, c_y) \leftarrow \frac{1}{|\mathcal{V}|}\sum_{{(x,y)}\in\mathcal{V}}(x,y)$ \hfill // Center x-coordinate and y-coordinate

\STATE
\STATE \textbf{Step 4: Compute Maximum Distance}
\FOR{each $(x,y) \in \mathcal{V}$}
    \STATE $r \leftarrow \sqrt{(x - c_x)^2 + (y - c_y)^2}$
    \STATE $r_{\text{max}} = \max_{(x,y) \in \mathcal{V}} \sqrt{(x-c_x)^2 + (y-c_y)^2}$
\ENDFOR

\STATE
\STATE \textbf{Step 5: Generate Colored Pixels with Relative Distance}
\FOR{each $(x,y) \in \mathcal{V}$}
    \STATE \textit{// Compute Relative Distance (darker at center, lighter at edges)}
    \STATE $d \leftarrow \sqrt{(x - c_x)^2 + (y - c_y)^2}$
    \STATE $\alpha_{\text{distance}} \leftarrow 1.0 - \gamma_{\text{distance}} \cdot \frac{d}{r_{\max}}$ \hfill // Relative Distance
    
    \STATE
    \STATE \textit{// Generate random noise}
    \STATE $\boldsymbol{\epsilon} \sim \mathcal{N}(\mathbf{0}, (\delta_{\text{noise}})^2 \cdot I)$ \hfill // Gaussian noise
    
    \STATE
    \STATE \textit{// Compute final color}
    \STATE $C^{(i)}_{\text{noise}} \leftarrow C^{(i)}_{\text{base}} + \sigma_c^{(i)} \odot \boldsymbol{\epsilon}$ \hfill // Add noise variation
    \STATE $C^{(i)}_{\text{final}} \leftarrow \alpha_{\text{distance}} \cdot C^{(i)}_{\text{noise}}$ \hfill //  Compute final color
    
    \STATE
    \STATE \textit{// Assign to output}
    \STATE $V_{\text{color}}^{(i)} \leftarrow C^{(i)}_{\text{final}}$
\ENDFOR

\STATE
\RETURN $V_{\text{color}}^{(i)}$
\end{algorithmic}
\end{algorithm}

Following this, we utilize the LaMa (Large Mask Inpainting) technique \cite{b19} to restore and refine the identified salient regions within the fundus images, generating inpainted images $I^{(i,t+1)} = \text{LaMa}(I^{(i,t)}, I_{\text{binary}}^{(i,t)}; \phi)$. LaMa is a state-of-the-art deep learning-based inpainting method that employs fast Fourier convolutions (FFCs) to effectively handle large irregular masks and generate high-quality reconstructions with global context awareness to ``erase'' these identified salient regions. The technique excels at preserving fine-grained details while maintaining global coherence, making it particularly suitable for medical images 
restoration tasks. This approach allows our model to iteratively focus on critical retinal areas that are essential for accurate DR grading. By intelligently reconstructing and enhancing the visibility of significant anatomical features, the inpainting process ensures that the model obtains a clearer and more comprehensive view of the retinal structures, which is crucial for robust clinical analysis and diagnostic accuracy.

After LaMa inpainting, the newly inpainted fundus images $I^{(i,t+1)}$ are reintroduced to the DR classification model to obtain updated predictions $\hat{y}^{(i,t+1)}$ and 
predicted classes $\hat{c}^{(i,t+1)}$. This iterative severity regression process continues as long as the prediction indicates a referable case ($\hat{c}^{(i,t)} \in \mathcal{C}_{\text{referable}}$), with each iteration identifying and removing additional lesion regions. The algorithm terminates when the inpainted fundus image is predicted as non-referable ($\hat{c}^{(i,t)} \in \mathcal{C}_{\text{non-referable}}$), indicating that the major lesions contributing to the high-severity classification have been successfully localized and removed. The total number of iterations $T_i$ required for each image $i$ is recorded. The final outputs---the fully inpainted 
images $\{I^{(i,T_i)}\}_{i=1}^{N_{\text{batch}}}$ and accumulated lesion masks $\{I_{\text{lesion}}^{(i,T_i)}\}_{i=1}^{N_{\text{batch}}}$---provide comprehensive lesion localization results for downstream diagnostic applications.

During the severity regression-based lesion localization process, the LaMa inpainting model may inadvertently remove or damage vessel structures within the identified lesion regions, leading to discontinuous or fragmented vessel appearances in the restored fundus images. To address this limitation and preserve the anatomical integrity of retinal vasculature, we propose a vessel enhancement and repair mechanism that operates on each inpainted image $I^{(i,t)}$ generated during the iterative process, as detailed in Algorithm~\ref{alg:vessel_repair}.

Firstly, in Phase 1, we utilize a pre-trained vessel segmentation model \cite{liu2024vsg} to obtain the predicted vessel mask $V_{\text{seg}}^{(i)}$ from the original fundus image $I^{(i,1)}$. This predicted vessel mask $V_{\text{seg}}^{(i)}$ are preprocessed through binarization (threshold=20), morphological operations (opening and closing) for noise removal, and filtering of short vessel segments (minimum length=20 pixels), which is denoted as $V^{(i)}_{\text{preprocess}}$. In Phase 2, a critical emphasis in vessel repair is identifying regions where the inpainted lesion areas intersect with vessel structures, as these regions are most susceptible to vessel damage during the inpainting process. In Phase 2, We detect the intersection regions at iteration $t$ by computing $V_{\text{inter}}^{(i,t)} = I_{\text{binary}}^{(i,t)} \cap V_{\text{preprocess}}^{(i)}$, where $I_{\text{binary}}^{(i,t)}$ represents the lesion mask from the iteration $t$.

In Phase 3, to ensure that the enhanced vessels maintain color consistency with the original fundus image, we extract vessel color statistics from the original image $I^{(i,1)}$. Specifically, we compute the mean color $\mu_c^{(i)}$ and standard deviation $\sigma_c^{(i)}$ by sampling pixel values in the vessel regions of $I^{(i,1)}$. These statistics are used to generate a colored vessel image $V_{\text{color}}^{(i,t)} = \text{GenerateColoredVessels}(V_{\text{preprocess}}^{(i)}, \mu_c^{(i)}, \sigma_c^{(i)})$ that matches the original vessel appearance. This vessel color generation process synthesizes realistic vessel appearances by incorporating spatial intensity variations and stochastic color perturbations, which is detailed in Algorithm \ref{alg:generate_colored_vessels}. During the vessel color generation process, given the preprocessed binary vessel mask $V_{\text{preprocess}}^{(i)}$ and extracted color statistics $\mu_c^{(i)}, \sigma_c^{(i)}$ from the original image $I^{(i,1)}$, we first extract the initial vessel coordinates $\mathcal{V} \leftarrow \{(x,y) \mid v_{xy} = 1, \text{ where } v_{xy} \triangleq V^{(i)}_{\text{preprocess}}[x, y] \}$. In step 2, we then compute a darkened base color 
$C^{(i)}_{\text{base}} = \beta_{\text{dark}} \cdot \mu_c^{(i)}$ to simulate the characteristic darker appearance of retinal vessels. Step 3 and step 4 determine the geometric centroid ($c_x, c_y$) of the vessel coordinates $\mathcal{V}$ and compute the maximum distance $r_{\text{max}} = \max_{(x,y) \in \mathcal{V}} \sqrt{(x-c_x)^2 + (y-c_y)^2}$ for normalization. In step 5, For each vessel pixel $(x,y)$, we generate its color through a three-stage process:
(1) computing the relative distance factor $\alpha_{\text{distance}} = 1.0 - \gamma_{\text{distance}} \cdot (\frac{d}{r_{\text{max}}})$, where $d = \sqrt{(x-c_x)^2 + (y-c_y)^2}$, to create a distance-based weighting scheme that is darker at the center and lighter at the periphery;
(2) sampling random noise ($\boldsymbol{\epsilon} \sim \mathcal{N}(\mathbf{0}, \delta_{\text{noise}}^2 \cdot I)$) and computing ($C^{(i)}_{\text{noise}} = C^{(i)}_{\text{base}} + \sigma_c^{(i)} \odot \boldsymbol{\epsilon}$) to model natural color variation; and
(3) obtaining the final color ($C^{(i)}_{\text{final}} = \alpha_{\text{distance}} \cdot C^{(i)}_{\text{noise}}$).
This distance-based weighting scheme mimics the physiological property that vessels appear darker near their origin at the optic disc, where they are thicker, gradually becoming lighter in peripheral regions where they branch and taper. The stochastic noise component enhances photorealism by introducing pixel-level color variation, reducing synthetic uniformity.
The algorithm operates with $(\mathcal{O}(|\mathcal{V}|)$) complexity, efficiently generating a colored vessel image ($V_{\text{color}}^{(i)}$) that preserves anatomical plausibility while facilitating seamless integration with inpainted regions in subsequent blending.

In Phase 4, the colored vessel image is then blended with the inpainted image $I^{(i,t)}$ using an adaptive blending strategy. For pixels in the intersection region $V_{\text{inter}}^{(i,t)}$, where vessel damage is most likely to occur, we apply an enhanced blending weight $\alpha_{\text{inter}}$ to strengthen vessel visibility: $I_{\text{enhance}}^{(i,t)}(x,y) = (1-\alpha_{\text{inter}}) I^{(i,t)}(x,y) + \alpha_{\text{inter}} V_{\text{color}}^{(i,t)}(x,y)$. For other vessel regions, we use a standard blending weight $\alpha_{\text{vessel}}$ to maintain natural appearance while preserving vessel structures. This adaptive blending approach ensures that vessels are appropriately enhanced in regions affected by inpainting while maintaining a natural appearance in unaffected areas, thereby preserving both the diagnostic utility and visual quality of the fundus images throughout the severity regression-based lesion localization process.

The vessel enhancement and repair mechanism operates seamlessly within the iterative framework, being applied to each inpainted image $I^{(i,t)}$ before it is fed back to the DR classification model $f_{\text{VL}}$ in the next iteration. This ensures that the model receives vessel-enhanced images that maintain anatomical integrity, preventing false predictions caused by vessel damage and improving the reliability of the severity regression-based lesion localization process.

Overall, this iterative refinement cycle, combining saliency map generation with inpainting and vessel repair, fosters a deeper understanding of complex visual patterns associated with various grades of DR.
The vessel repair process specifically addresses the challenge of maintaining vascular continuity in lesion-affected regions, ensuring that the model learns from anatomically complete retinal structures rather than fragmented vessel networks.
By continuously updating the model's focus based on the insights gained from previous iterations—including both lesion context and preserved vessel topology—we improve its diagnostic performance and capability to interpret intricate clinical information from retinal images, enabling more accurate assessment of DR severity while maintaining the structural integrity of the underlying vascular architecture.

\section{Experimental Setup}
\begin{table*}[htbp]
\centering
\caption{The distribution of disease and lesion annotations in the three datasets. Diseases labels consist of no DR, mild DR, moderate DR, severe DR and Proliferable DR. Lesion labels consists of hard exudates (EX), hemorrhages (HE), microaneurysm (MA), soft exudates (SE).}
\label{tab:distribution}
\scalebox{0.85}{ 
\begin{tabular}{c c c c c c c c c c c}
\hline
\multirow{2}{*}{Dataset} & \multirow{2}{*}{Total} & \multicolumn{5}{c}{Disease Labels} & \multicolumn{4}{c}{Lesion Labels} \\
\cline{3-11}
& & No DR & Mild & Moderate & Severe & Proliferative & EX & HE & MA & SE \\
\hline
FGADR & 1842 & 101 & 213 & 594 & 647 & 287 & 1279 & 1456 & 1424 & 627 \\
DDR & 1510 & 755 & 98 & 548 & 34 & 74 & 485 & 599 & 569 & 238 \\
PrivateData & 235 & 23 & 128 & 58 & 16 & 10 & 132 & 135 & 206 & 62 \\
IDRiD & 516 & 168 & 25 & 168 & 93 & 62 & - & - & - & -  \\
\hline

\end{tabular}
}
\end{table*}

\subsection{Dataset}
In our experimental setup, we have four datasets, including the DDR dataset \cite{li2019diagnostic}, the FGADR dataset \cite{zhou2021benchmark}, the IDRiD dataset \cite{porwal2018indian}, and our private dataset. The details are illustrated in Table \ref{tab:distribution}. All three datasets are used for the DR grading task, and they are annotated with five DR grades and four types of retina lesions. The DDR dataset comprises a total of 13,673 fundus images, of which we focus on a subset containing 757 images that include both lesion classifications and DR grading labels. This selection is crucial as it provides the necessary annotations for our multi-modal learning tasks. To ensure a balanced representation of classes in our experiments, we also incorporate an additional 755 healthy images, resulting in a refined DDR-subset dataset. 

\subsection{Implementation Details}
All experiments are conducted on an NVIDIA RTX 4090 GPU. We resize input images to 384 × 384 pixels for our models. For the DR grading and lesion classification task, we employ CaiT-xs24 and ClinicalBERT as the vision and text backbones, respectively. An AdamW optimizer is used with LinearWarmupCosineAnnealingLR as the learning rate scheduler. Both the learning rate and weight decay are set to 1e-5. The model is trained for 200 epochs with a batch size of 16. Data augmentation techniques, including rotation, horizontal flipping, and color jittering, are applied to the training images. Additionally, CLAHE (Contrast Limited Adaptive Histogram Equalization) is applied to enhance the contrast of fundus images. All datasets are stratified and split into training and validation sets.

\subsection{Evaluation Metrics}
To evaluate the performance of the DR grading task, since this is a single-label classification task, we utilize area-under-the-curve (AUC), quadratic weighted kappa (Kappa), sensitivity (Sens), and specificity (Spec) as evaluation metrics. For the joint DR grading and lesion classification, which is a multi-label classification task, we employ accuracy, AUC, and F1-score as the evaluation metrics.

\subsection{Baseline}
\textbf{ResNet101} \cite{he2016deep} is a widely recognized CNN architecture that serves as a backbone for various image classification tasks.

\textbf{ViT-B/16} \cite{dosovitskiy2020image} is a ViT model that processes images by dividing them into 16 $\times $16 pixel patches and applying a transformer architecture for image classification tasks.

\textbf{CaiT-xs24} \cite{wen2024concept} is a ViT variant that combines convolutional layers with transformer mechanisms to enhance feature extraction and capture global context.

\textbf{CABNet} \cite{b22} is designed to detect DR diagnosis by integrating a category attention block with convolutional layers.

\textbf{CANet} \cite{b25} utilizes specific disease attention to adaptively recalibrate feature responses, enhancing the model's ability to perform DR and DME detection.

\textbf{LbCL} \cite{huang2021lesion} leverages contrastive learning techniques to improve feature discrimination, making it suitable for multi-class DR classification tasks.

\textbf{MIL-VT} \cite{yu2021mil} applies multiple instance learning principles within a ViT framework. This model effectively captures intricate patterns within images by considering multiple regions, making it highly effective for identifying lesions and grading the severity of DR.

\textbf{CLAT} \cite{wen2024concept} incorporates attention mechanisms within a contrastive learning framework to enhance the model's focus on relevant features while learning from both positive and negative examples. This approach aids in improving feature representation and classification accuracy within the context of DR tasks.

\textbf{CLIP-DR} \cite{yu2024clip} leverages contrastive language-image pretraining (CLIP) to enhance DR analysis. By aligning visual features with clinical text descriptors, this model improves lesion interpretation and severity grading, offering a unified framework for robust and explainable DR diagnosis.

\begin{table*}[!t]
    \centering
    \caption{DR Grading Performance metrics of various models across Private Dataset, DDR-subset Dataset, and IDRiD dataset. The results are the mean values of the 10-fold cross-validation (unit: \%).  \textbf{BOLD} indicates the optimal performance.}
    \label{table1:DR performance}
    \scalebox{0.85}{ 
    \begin{tabular}{lcccccccccccc}
        \toprule
        \multirow{2}{*}{Models} & \multicolumn{4}{c}{Private Data} & \multicolumn{4}{c}{DDR-subset} & \multicolumn{4}{c}{IDRiD} \\
        \cmidrule(lr){2-5} \cmidrule(lr){6-9} \cmidrule(lr){10-13}
        & AUC & kappa & Sen & Spec & AUC & kappa & Sen & Spec & AUC & kappa & Sen & Spec \\
        \midrule
        ResNet101 & 88.56 & 76.43 & 81.27 & 91.83 & 89.03 & 75.08 & 82.01 & 92.07 & 87.39 & 82.54 & 81.12 & 93.21 \\
        ViT-B/16 & 87.92 & 78.12 & 81.45 & 92.76 & 87.15 & 79.03 & 82.14 & 93.15 & 86.14 & 83.24 & 81.10 & 90.21 \\
        CaiT-xs24 & 89.14 & 84.37 & 82.91 & 94.23 & 88.76 & 85.26 & 83.12 & 95.11 & 87.31 & 83.83 & 81.42 & 92.31 \\
        \midrule
        CABNet & 89.02 & 85.73 & 82.65 & 93.42 & 88.23 & 86.25 & 83.17 & 93.17 & 85.26 & 85.89 & 81.17 & 92.17 \\
        CANet & 86.98 & 84.92 & 82.37 & 91.86 & 87.72 & 86.21 & 83.11 & 92.11 & 82.12 & 84.14 & 80.28 & 93.11 \\
        LbCL & 89.37 & 85.83 & 82.91 & 93.12 & 88.76 & 85.14 & 83.12 & 92.29 & 88.13 & 84.10 & 81.18 & 93.14 \\
        MIL-VT & 90.85 & 87.36 & 83.42 & 92.91 & \textbf{91.72} & 87.14 & 82.12 & 90.16 & 87.12 & 86.41 & 82.14 & 91.12 \\
        CLAT & 89.93 & 88.15 & 84.21 & 95.46 & 86.68 & 88.66 & 84.93 & 96.23 & 89.12 & 85.12 & 83.10 & 92.12 \\
        CLIP-DR & 90.62 & 89.47 & 84.76 & 96.32 & 88.12 & 89.14 & 83.14 & 96.12 & 89.42 & 85.26 & 81.24 & 90.14 \\
        \midrule
        \textbf{Ours} & \textbf{92.18} & \textbf{89.53} & \textbf{85.29} & \textbf{97.14} & 91.44 & \textbf{89.61} & \textbf{85.93} & \textbf{97.95} & \textbf{90.79} & \textbf{86.71} & \textbf{83.16} & \textbf{94.17} \\
        \bottomrule
    \end{tabular}
    }
\end{table*}

\begin{table*}[!t]
    \centering
    \caption{DR and Lesion Detection Performance of various models on Private dataset, DDR-subset dataset, and FGADR dataset. Results are averaged over 10-fold cross-validation (unit: \%). \textbf{BOLD} indicates the optimal performance.}
    \label{table2:DR and lesion performance}
 \scalebox{0.85}{ 
    \begin{tabular}{lcccccccccc}
        \toprule
        \multirow{2}{*}{Models} & \multicolumn{3}{c}{Private Data} & \multicolumn{3}{c}{DDR-subset} & \multicolumn{3}{c}{FGADR} \\
        \cmidrule(lr){2-4} \cmidrule(lr){5-7} \cmidrule(lr){8-10}
        & ACC & AUC & F1 & ACC & AUC & F1 & ACC & AUC & F1 \\
        \midrule
        ViT-B/16 & 85.95 & 90.83 & 84.79 & 88.33 & 91.04 & 77.86 & 84.88 & 90.72 & 80.35 \\
        ViT-B/32 & 84.98 & 91.85 & 77.92 & 87.23 & 92.74 & 78.72 & 83.40 & 90.70 & 79.44 \\
        ViT-L/16 & 84.93 & 90.19 & 70.82 & 85.32 & 90.76 & 71.67 & 84.10 & 92.66 & 81.06 \\
        ViT-L/32 & 85.04 & 90.97 & 71.58 & 85.42 & 91.36 & 72.36 & 84.12 & 90.95 & 80.97 \\
        CaiT-xs24 & 85.92 & 92.36 & 74.69 & 86.57 & 92.84 & 75.21 & 84.52 & 93.60 & 83.74 \\
        \midrule
        CT & 78.86 & 68.75 & 62.97 & 79.43 & 69.29 & 63.56 & 75.07 & 68.18 & 79.94 \\
        PCBM & 82.63 & 89.74 & 75.48 & 83.07 & 90.19 & 76.31 & 55.66 & 51.23 & 64.64 \\
        CLAT & 83.78 & 93.85 & 74.67 & 93.14 & 94.02 & \textbf{85.19} & 82.66 & 82.86 & 87.57 \\
        CLIP-DR & 86.28 & 94.06 & 78.38 & 92.87 & 93.65 & 84.94 & 86.53 & 93.18 & 82.76 \\
        \midrule
        \textbf{Ours} & \textbf{89.08} & \textbf{95.21} & \textbf{84.36} & \textbf{93.64} & \textbf{94.75} & 83.15 & \textbf{87.61} & \textbf{94.32} & \textbf{89.98} \\
        \bottomrule
    \end{tabular}
    }
\end{table*}

\begin{table*}[!t]
    \centering
    \caption{Performance of different vision backbones in TWLR for joint DR grading and lesion classification on private dataset, DDR-subset and FGADR dataset.  All models use ClinicalBERT as the text encoder. Results are averaged over 10-fold cross-validation (unit: \%).}
    \label{table3:different backbone}
    \scalebox{0.85}{ 
    \begin{tabular}{lcccccccccc}
        \toprule
        \multirow{2}{*}{Vision Backbone} & \multicolumn{3}{c}{Private Data} & \multicolumn{3}{c}{DDR-subset} & \multicolumn{3}{c}{FGADR} \\
        \cmidrule(lr){2-4} \cmidrule(lr){5-7} \cmidrule(lr){8-10}
        & ACC & AUC & F1 & ACC & AUC & F1 & ACC & AUC & F1 \\
        \midrule
        ViT-B/16 & 82.97 & 92.34 & 73.33 & 79.40 & 89.64 & 66.06 & 84.19 & 92.63 & 81.23 \\
        ViT-B/32 & 83.45 & 92.35 & 74.35 & 82.04 & 89.59 & 69.49 & 82.20 & 91.53 & 79.25 \\
        ViT-L/16 & 83.45 & 92.35 & 74.07 & 86.16 & 91.03 & 73.78 & 83.50 & 92.83 & 80.58 \\
        ViT-L/32 & 84.16 & 92.20 & 75.45 & 85.21 & 90.55 & 70.85 & 81.36 & 91.89 & 81.14 \\
        \midrule
        CaiT-xs24 & \textbf{86.28} & \textbf{94.06} & \textbf{78.32} & \textbf{88.52} & \textbf{94.75} & \textbf{78.96} & \textbf{85.94} & \textbf{94.32} & \textbf{83.41} \\
        \bottomrule
    \end{tabular}
    }
\end{table*}

\subsection{Experimental Results on Classification}
\subsubsection{DR Classification Performance}
In Table~\ref{table1:DR performance}, the top three models, ResNet101, ViT-B/16, and CaiT-xs24, serve as baselines, with ViT-B/16 and CaiT-xs24 being variants of the ViT architecture, leveraging self-attention mechanisms for feature extraction. The subsequent models, LBCL and MIL-VT, are specialized classifiers known for their strong performance in DR grading, employing contrastive learning and ViT-based attention mechanisms, respectively, to effectively capture DR-related features. Additionally, CABNet and CANet are attention-based models specifically designed to address DR-related challenges, with CABNet focusing on handling class imbalance through category-specific attention blocks, and CANet enabling joint grading of DR and diabetic macular edema via cross-disease attention mechanisms. Furthermore, CLAT and CLIP-DR are vision-language models tailored for DR tasks, incorporating expert knowledge from textual data to enhance their grading capabilities. Despite this, our proposed model demonstrates superior performance across various metrics on the Private, DDR-subset, and IDRiD datasets. This advantage stems from our effective integration of textual clinical ophthalmological knowledge, which explicitly captures the intrinsic relationship between lesion characteristics and DR severity levels, enabling the model to better understand how distinct DR grades correspond to specific lesion patterns and thereby improve grading accuracy.

\subsubsection{DR and Lesion Classification Performance}
In our experimental analysis of combined DR grading and lesion classification, as presented in Table~\ref{table2:DR and lesion performance}, our proposed model demonstrates superior performance across multiple datasets. On the private dataset, our model significantly outperforms baselines such as ViT-B/16 and ViT-L/32 in terms of accuracy, AUC, and F1 score, showing notable improvements in classification consistency. Compared to other models like CaiT-xs24 and PCBM, our approach achieves a better balance in lesion classification and grading, particularly with a substantial increase in F1 score on the DDR-subset and FGADR datasets. Against vision-language models like CLAT and CLIP-DR, where the latter may utilize some strategies to adjust the semantic space among different DR categories, our model maintains a competitive edge, especially in AUC on the DDR-subset and F1 on FGADR. These results highlight the robustness of our model in effectively addressing the challenges of joint DR grading and lesion classification across diverse datasets.

\begin{table*}[!t]
  \centering
  \caption{Lesion Pixel Distribution Analysis. ``Average Proportion'' indicates the mean lesion area as a percentage of total image size, with most lesions occupying less than 1\% of the fundus image. The remaining columns show image counts within specified pixel ranges.}
  \label{tab:lesion_analysis}
    \scalebox{0.85}{ 
  \begin{tabular}{c c c c c c c c c c}
    \toprule
    \multirow{2}{*}{Dataset} & \multirow{2}{*}{Lesion} & Average & \multicolumn{7}{c}{Count Range (\# lesion pixels)} \\
    \cmidrule(lr){4-10}
     & & Proportion (\%) & 1-10 & 10-100 & 100-500 & 500-1K & 1K-10K & 10K-100K & 100K+ \\
    \midrule
    \multirow{6}{*}{FGADR} & MA & 0.38 & 91 & 630 & 443 & 78 & 177 & 4 & 0 \\
     & EX & 1.05 & 27 & 206 & 387 & 205 & 419 & 34 & 0 \\
     & HE & 1.41 & 7 & 111 & 385 & 267 & 641 & 45 & 0 \\
     & SE & 0.47 & 4 & 83 & 282 & 135 & 123 & 0 & 0 \\
    \noalign{\smallskip}
    \cline{2-10} 
    \noalign{\medskip}
     & Overall & 3.31 & 2 & 109 & 511 & 439 & 2068 & 1428 & 107 \\
    \midrule
    \multirow{6}{*}{DDR} & MA & 0.02 & 157 & 367 & 40 & 2 & 0 & 0 & 0 \\
     & EX & 0.29 & 27 & 206 & 155 & 47 & 49 & 0 & 0 \\
     & HE & 0.44 & 21 & 225 & 201 & 74 & 74 & 6 & 0 \\
     & SE & 0.19 & 81 & 129 & 17 & 12 & 0 & 0 & 0 \\
    \noalign{\smallskip}
    \cline{2-10} 
    \noalign{\medskip}
     & Overall & 0.94 & 676 & 48 & 274 & 266 & 900 & 362 & 44 \\
    \bottomrule
  \end{tabular}
  }
\end{table*}

\begin{table*}[htbp]
  \centering
  \caption{WEAKLY SUPERVISED SEGMENTATION PERFORMANCE BASED ON DIFFERENT LESIONS ON FGADR AND DDR DATASETS USING OUR TWLR METHOD (UNIT: \%)}
  \label{tab:segmentation_performance}
  \scalebox{0.85}{  
  \begin{tabular}{c l c c c c c c c}
    \toprule
    \multirow{2}{*}{Dataset} & \multirow{2}{*}{Metric} & \multirow{2}{*}{bg} & \multirow{2}{*}{MA} & \multirow{2}{*}{EX} & \multirow{2}{*}{HE} & \multirow{2}{*}{SE} & \multirow{2}{*}{Overall(w/o bg)} & \multirow{2}{*}{Overall} \\
    & & & & & & & & \\
    \midrule
    \multirow{3}{*}{FGADR} & sensitivity & 95.46 & 66.6 & 92.4 & 81.6 & 94.3 & 83.7 & 86.07 \\
    & mIoU & 93.49 & 2.0 & 5.6 & 5.2 & 4.1 & 4.15 & 22.02 \\
    & Dice & 94.34 & 2.2 & 8.6 & 7.3 & 4.8 & 5.73 & 23.45 \\
    \midrule
    \multirow{3}{*}{DDR} & sensitivity & 96.48 & 47.2 & 86.1 & 53.8 & 89.5 & 69.15 & 74.62 \\
    & mIoU & 93.78 & 3.1 & 4.6 & 5.2 & 4.9 & 4.45 & 22.31 \\
    & Dice & 93.24 & 3.2 & 4.7 & 4.7 & 5.7 & 4.58 & 22.25 \\
    \bottomrule
  \end{tabular}
  }
\end{table*}

\subsubsection{Evaluation of Vision Backbones for Joint DR Grading and Lesion Classification in TWLR} To further investigate the efficacy of our vision-language model (TWLR) in the joint task of DR grading and lesion classification, we evaluated the impact of Lesion Classification of various vision backbones, as detailed in Table~\ref {table3:different backbone}. Our analysis reveals a clear trend in performance across the ViT and its variants on the Private, DDR-subset, and FGADR datasets. Among the ViT-based models, including ViT-B/16, ViT-B/32, ViT-L/16, and ViT-L/32, the CaiT-xs24 variant consistently demonstrates the most robust adaptability and superior performance across all datasets. Specifically, CaiT-xs24 exhibits marked improvements in accuracy, AUC, and F1 scores, particularly on the DDR-subset and FGADR datasets, where it achieves the best balance between grading precision and lesion classification capability. In contrast, other ViT models, such as ViT-B/16 and ViT-L/32, show relatively modest performance, with noticeable gaps in lesion classification sensitivity. These findings underscore the suitability of CaiT-xs24 for the complex demands of joint DR grading and lesion classification, owing to its enhanced feature extraction capabilities and architectural efficiency. Consequently, we selected CaiT-xs24 as the vision backbone for the vision component of our TWLR model, ensuring optimal performance in this multifaceted task.

\subsection{Weakly Localization Performance}

\subsubsection{Overall Weakly Semantic Segmentation Performance on Lesions}
The extreme scarcity of lesion regions in fundus images presents a fundamental challenge for weakly supervised quantification. As shown in Table \ref{tab:lesion_analysis}, diabetic retinopathy lesions occupy remarkably small proportions of fundus images, with overall lesion coverage averaging only 0.59\% in FGADR and 0.15\% in DDR datasets. Individual lesion types exhibit even more severe imbalance, particularly microaneurysms (MA), which constitute merely 0.02\% of the image area in OIA-DDR. This extreme foreground-background imbalance poses significant challenges for weakly supervised methods that rely on class activation maps or attention mechanisms, which typically struggle to capture such fine-grained details without pixel-level supervision. Consequently, despite the clinical importance of accurate lesion quantification for DR grading and progression monitoring, the literature remains notably sparse on weakly supervised approaches for this task. Most existing works either resort to fully supervised segmentation with expensive pixel-level annotations or limit themselves to image-level classification without attempting precise lesion localization and quantification. The inherent difficulty of detecting and quantifying such minute pathological regions using only image-level labels explains why weakly supervised lesion quantification remains an underexplored yet critical challenge in medical image analysis.

Table \ref{tab:segmentation_performance} demonstrates the performance of our TWLR method for weakly supervised lesion segmentation across the FGADR and DDR datasets. The results reveal distinct performance patterns across different lesion types and evaluation metrics. For sensitivity, exudates (EX) consistently achieve the highest detection rates, exceeding 80\% on both datasets, while microaneurysms (MA) present the greatest challenge with detection rates below 50\% on the DDR dataset. Hard exudates and soft exudates show intermediate performance, with values ranging from 50\% to 85\% across datasets. In contrast, mIoU and Dice coefficients remain consistently modest across all lesion categories, with most values falling below 10\% when background is excluded. This pattern reflects the inherent challenges of retinal lesion segmentation, where lesions typically occupy less than 1\% of total image pixels, creating extreme class imbalance. The sparse distribution, combined with lesion morphological heterogeneity across different categories (MA, EX, HE, SE) and the limitations of weakly supervised boundary-level annotation, makes precise spatial overlap metrics inherently challenging. The performance gap between sensitivity and overlap-based metrics highlights the fundamental difficulty of precise lesion boundary delineation in tasks with extreme spatial sparsity, where small prediction errors significantly impact overlap-based evaluation measures. These performance patterns are consistent across both datasets, validating the robustness of our approach while reflecting the inherent complexity of retinal lesion segmentation.

\begin{figure*}[htp]
    \centering
    \begin{minipage}[t]{0.48\textwidth}
        \centering
        \includegraphics[width=\textwidth]{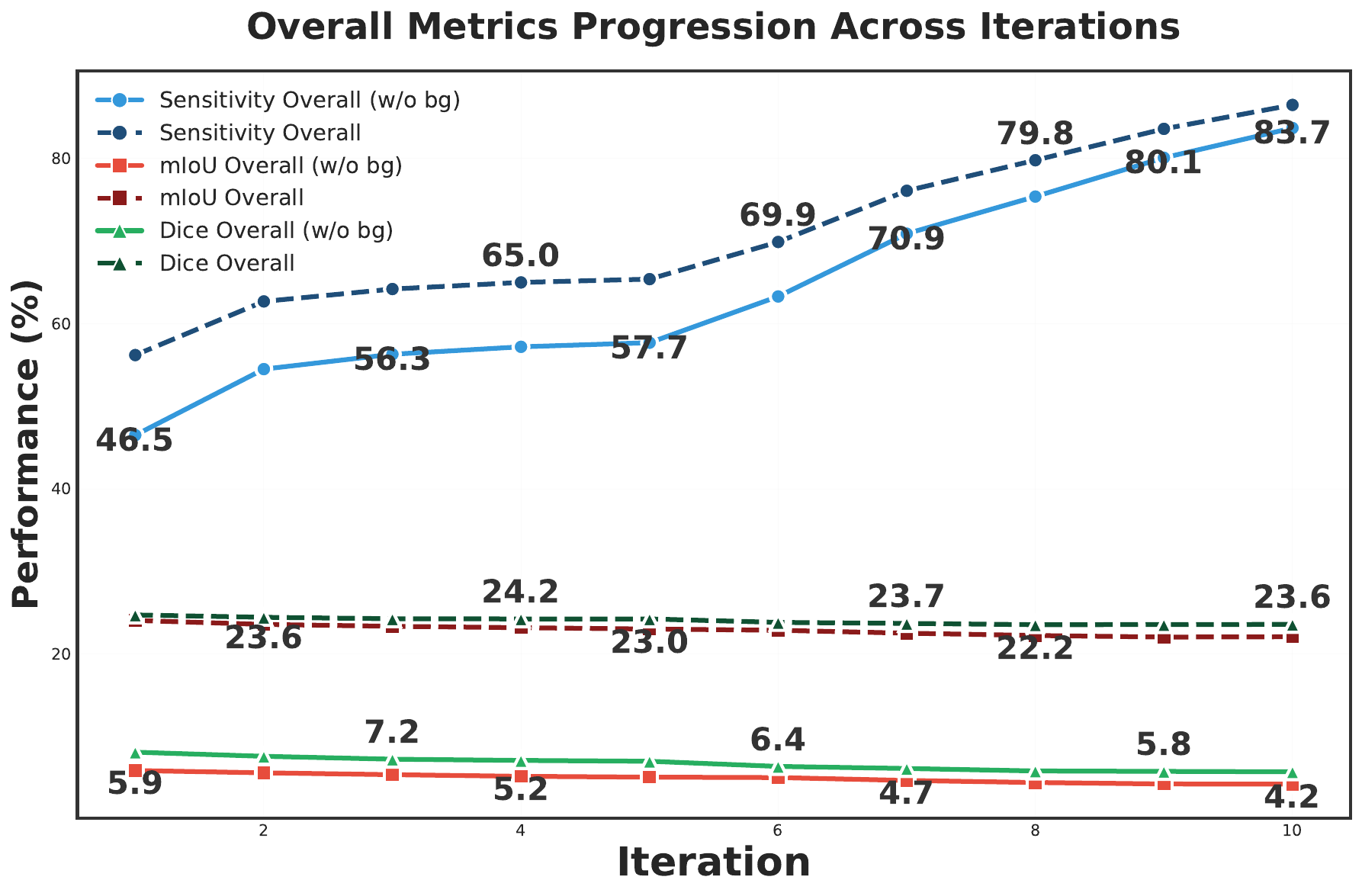}
        \captionof{figure}{Performance progression of the TWLR method across 10 lesion location iterations, showing sensitivity, mIoU, and Dice metrics across FGADR and DDR datasets.}
        \label{fig:iterative_metric}
    \end{minipage}
    \hfill
    \begin{minipage}[t]{0.48\textwidth}
        \centering
        \includegraphics[width=\textwidth]{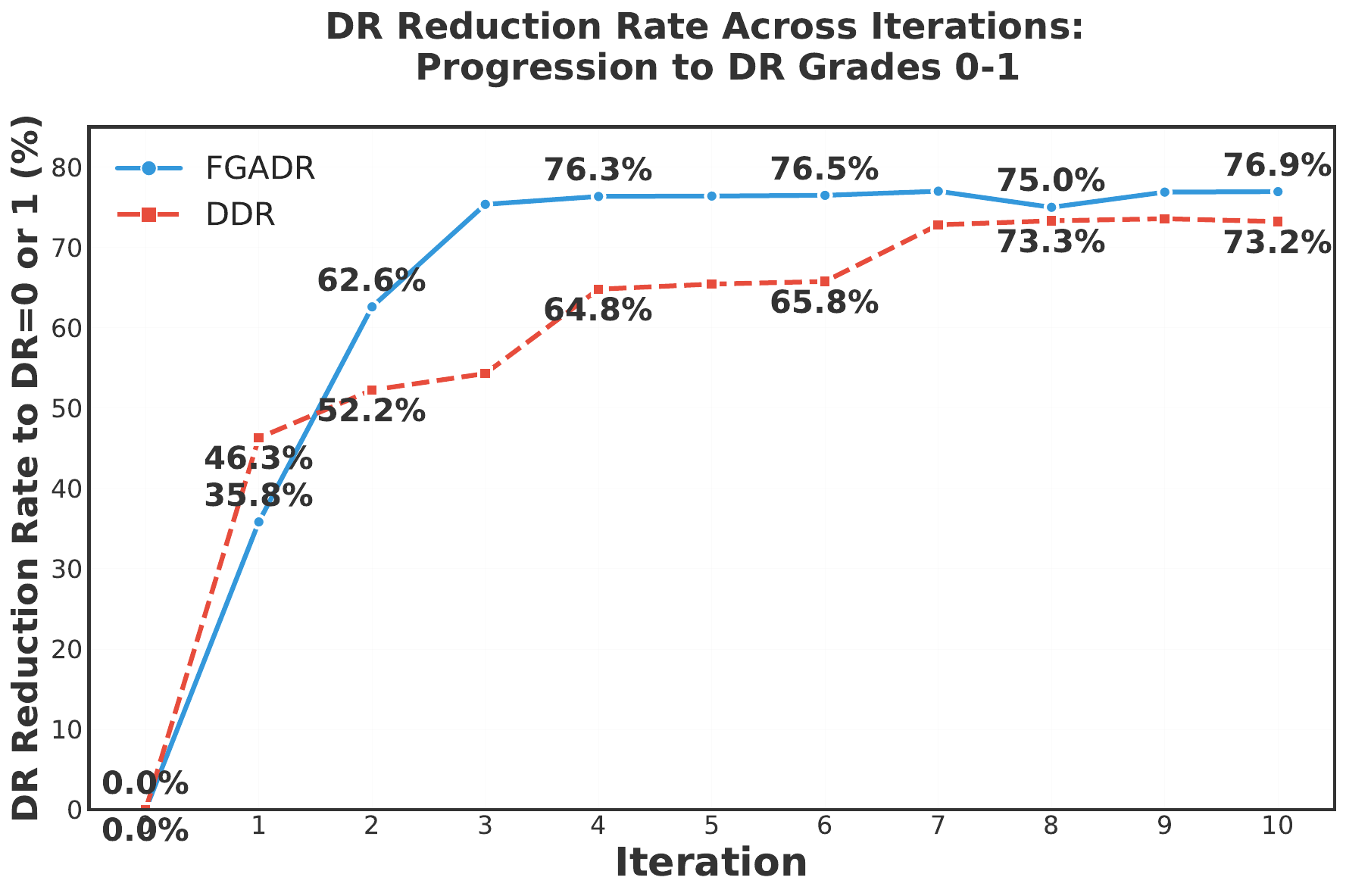}
        \captionof{figure}{DR Reduction Rate across iteration times. As iteration time increases, more severe DR fundus images are inpainted to healthier moderate DR status.}
        \label{fig:dr_reduction_rate}
    \end{minipage}
\end{figure*}

\subsubsection{Iterative Weakly Semantic Segmentation Performance on Lesions}
Fig. \ref{fig:iterative_metric} illustrates the performance evolution of our TWLR method across 10 lesion localization iterations, revealing the inherent trade-off between lesion classification capability and segmentation precision. The sensitivity metrics demonstrate substantial improvement throughout the iterative process, with overall sensitivity increasing from 56.3\% at iteration 1 to 83.7\% at iteration 10, while sensitivity excluding background progresses from 46.5\% to 80.1\%. This pronounced enhancement indicates that the method successfully identifies an increasing number of lesions as iterations advance, reflecting improved lesion classification capability.

\begin{figure*}[htbp]
    \centering
    \begin{minipage}[t]{0.48\textwidth}
        \centering
        \includegraphics[width=\textwidth]{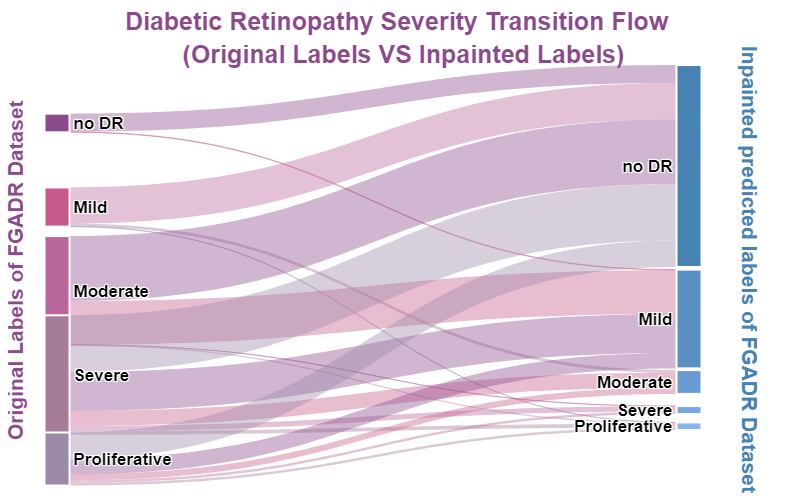}
        \captionof{figure}{Transition flow of DR severity grades between original labels and inpainted predicted labels in the FGADR dataset. Original DR labels (left) predominantly flow toward restored healthy states (right), with most diseased fundus images (DR grades 2-4) successfully transitioning to no DR (grade 0) or mild DR (grade 1), demonstrating effective lesion feature restoration and therapeutic potential of the iterative inpainting process.}
        \label{fig:transition_flow_fgadr}
    \end{minipage}
    \hfill
    \begin{minipage}[t]{0.48\textwidth}
        \centering
        \includegraphics[width=\textwidth]{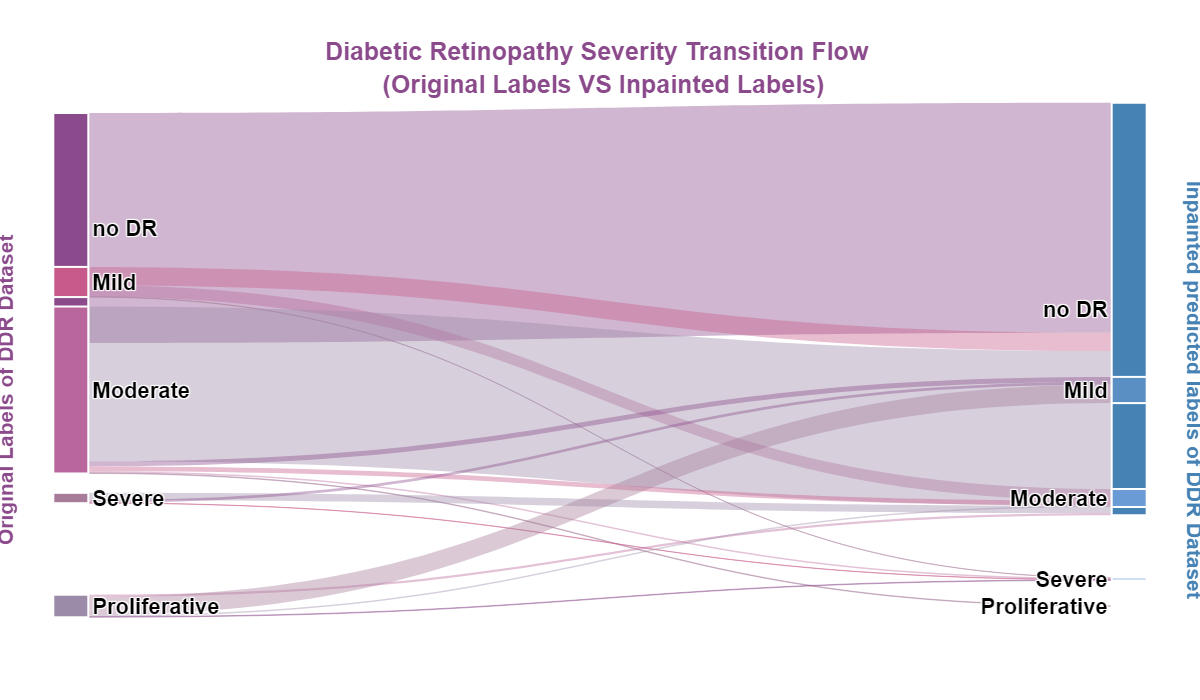}
        \captionof{figure}{Transition flow of DR severity grades between original labels and inpainted predicted labels in the DDR dataset. Original DR labels (left) predominantly flow toward restored healthy states (right), with most diseased fundus images (DR grades 2-4) successfully transitioning to no DR (grade 0) or mild DR (grade 1), demonstrating effective lesion feature restoration and therapeutic potential of the iterative inpainting process.}
        \label{fig:transition_flow_ddr}
    \end{minipage}
\end{figure*}

\begin{figure*}[htbp]
    \centering
    \begin{minipage}[t]{0.51\textwidth}
        \centering
        \includegraphics[width=\textwidth,height=0.6\textheight,keepaspectratio]{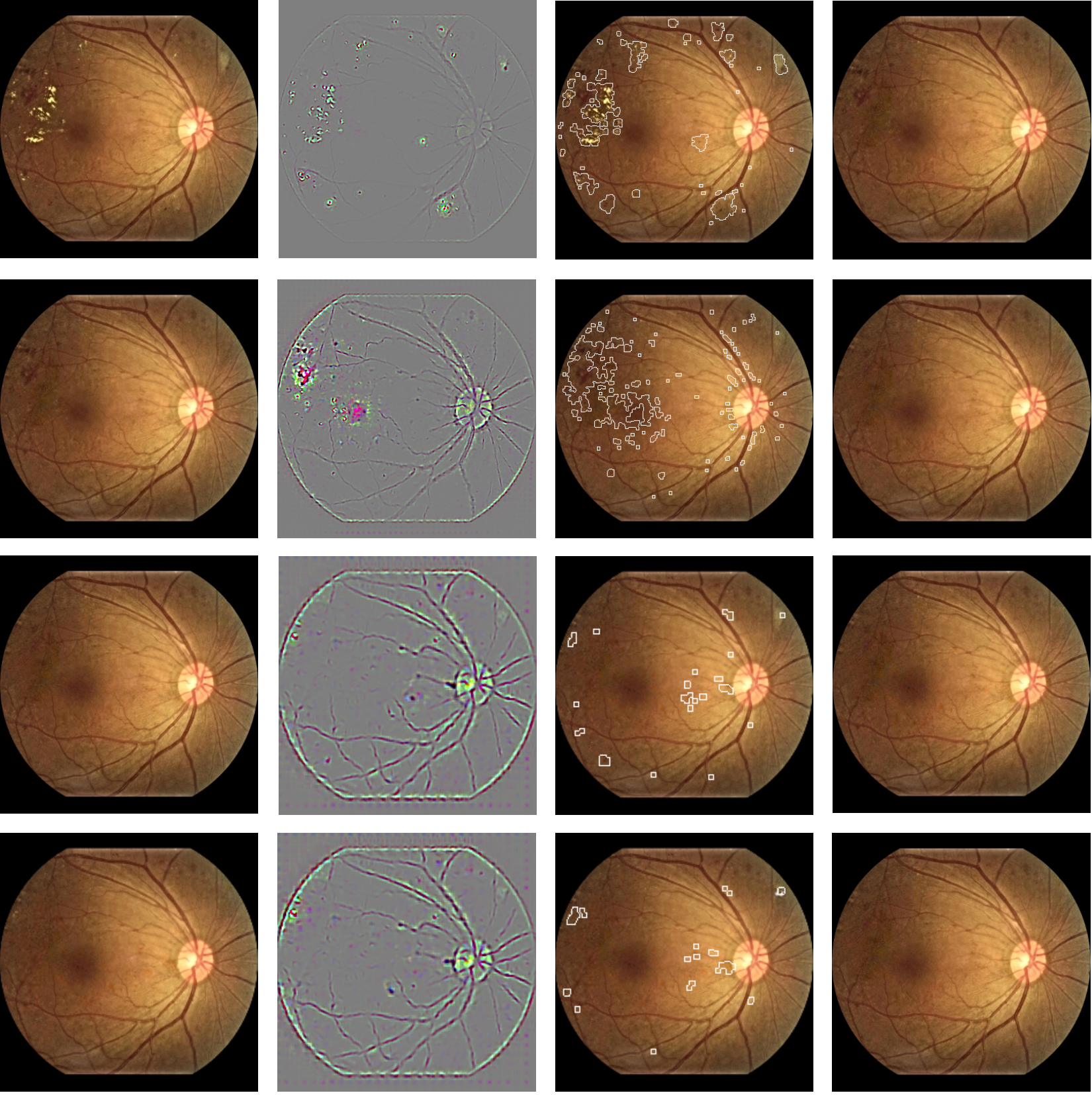}
        \captionof{figure}{Qualitative result on progression of the TWLR method across 1-4 lesion iterations on weakly supervised lesion segmentation tasks. Each row represents one iteration process. The first column is original image, second column is the saliency map extracted from the DR and lesion classification. The third column is the prediction contours in white obtained from the saliency map. The final column illustrates the fundus images inpainted based on the found lesion contours.}
        \label{fig:FGADR_0015_2}
    \end{minipage}
    \hfill
    \begin{minipage}[t]{0.40\textwidth}
        \centering
        \includegraphics[width=\textwidth,height=0.6\textheight,keepaspectratio]{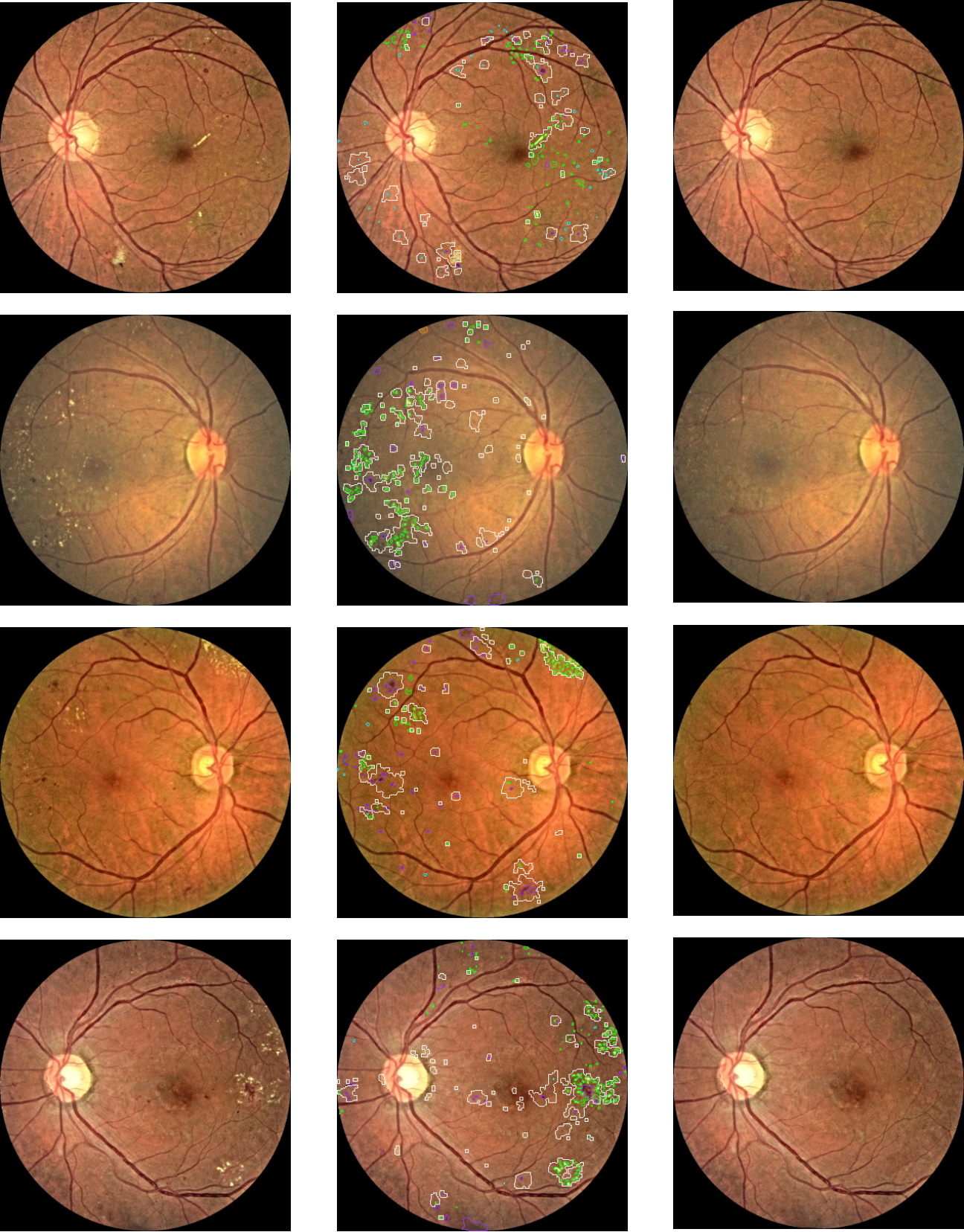}
        \captionof{figure}{Inpainted images that are predicted as no DR after iterations. The first row is original image, second row illustrates the original images, lesion prediction (white) with ground truth (green HE, purple EX, light-blue MA, orange SE). Third row is the inpainted images that are considered as no DR status.}
        \label{fig:origin_inpaint}
    \end{minipage}
\end{figure*}

However, this improved detection performance comes with a corresponding decrease in spatial precision metrics. The mIoU coefficients show a gradual decline from 7.2\% to 4.2\% (excluding background) and from 24.2\% to 23.6\% (overall), while Dice coefficients similarly decrease from 5.9\% to 4.2\% (excluding background) and remain relatively stable around 23\% (overall). This inverse relationship between sensitivity and overlap-based metrics reflects the fundamental challenge in lesion localization: as the method becomes more sensitive to potential lesion regions, it inevitably introduces more false positive detections, leading to reduced spatial overlap accuracy. The observed pattern demonstrates the classic precision-recall trade-off inherent in medical image analysis, where increased sensitivity for lesion classification results in more liberal classification boundaries, consequently reducing the precision of spatial localization. While the method successfully identifies a greater proportion of actual lesions across iterations, the accompanying increase in false positives manifests as diminished IoU and Dice scores. This behavior is characteristic of iterative localization approaches and highlights the need for careful balance between detection sensitivity and spatial precision in weakly supervised lesion segmentation tasks.

Fig. \ref{fig:dr_reduction_rate} demonstrates the therapeutic efficacy of our treatment approach across iterative interventions, measured by the progression of severe DR cases to healthier DR grades (0-1). The results reveal substantial improvement in both datasets, with FGADR showing a remarkable reduction rate progression from 0.0\% at baseline to 76.9\% after 10 iterations, while DDR demonstrates a parallel improvement trajectory from 0.0\% to 73.2\%. The most dramatic improvements occur within the initial iterations, with FGADR achieving a 62.6\% reduction by iteration 2 and DDR reaching 52.2\% at the same time point, indicating rapid therapeutic response in the early treatment phases.
The convergence patterns differ between datasets, with FGADR reaching a plateau around 76-77\% after iteration 3, maintaining relatively stable reduction rates between 75.0\% and 76.9\% through iterations 6-10. In contrast, DDR exhibits more gradual improvement, progressively advancing from 64.8\% at iteration 4 to 73.3\% at iteration 8, before stabilizing around 73.2\% by iteration 10. This difference may reflect varying baseline severity distributions or dataset-specific characteristics affecting treatment responsiveness.
The sustained high reduction rates achieved by both datasets ($>$73\%) demonstrate the method's effectiveness in converting severe DR cases to milder grades, representing clinically meaningful therapeutic outcomes. The consistent performance across different datasets validates the robustness of the approach, while the plateau effect observed in later iterations suggests optimal treatment dosing and timing. These findings indicate that the majority of treatable severe DR cases can be successfully managed to achieve healthier retinal status, with most therapeutic benefit realized within the first few intervention cycles.

\subsubsection{Qualitative Results}
\paragraph{Iterative progression on lesion segmentation}

Fig. \ref{fig:FGADR_0015_2} presents the qualitative progression of the TWLR method across 10 lesion localization iterations, demonstrating the sequential discovery and refinement of different retinal lesions in weakly supervised lesion segmentation tasks. Each row represents a different patient case, with the progression from left to right showing: (1) original fundus image, (2) intermediate segmentation results with detected lesions highlighted in different colors, (3) refined lesion classification, and (4) final segmentation output.
The iterative process reveals a hierarchical lesion classification pattern where the algorithm initially identifies the most prominent and well-defined lesion—primarily hard exudates (HE)—which appear as bright, well-circumscribed yellowish deposits. These lesions are typically easier to detect due to their high contrast against the retinal background and distinct morphological characteristics. As iterations progress, the method expands its detection capability to identify more subtle and morphologically diverse lesions, including microaneurysms (MA), soft exudates (SE), and hemorrhages (EX).
The color-coded visualization in the intermediate results (second column) illustrates the spatial distribution and progression of lesion classification, with different colors likely representing different lesion categories or confidence levels. The sequential improvement from initial detection to final output demonstrates the method's ability to progressively refine lesion boundaries and reduce false positives while maintaining sensitivity to true lesion features.
This qualitative analysis validates the quantitative trends observed in previous figures, showing how the iterative approach successfully expands lesion coverage while dealing with the inherent trade-off between sensitivity and precision in weakly supervised scenarios. The progression from detecting obvious lesions to more subtle lesions reflects the clinical relevance of the approach, as it mirrors the typical diagnostic process where prominent features are identified first, followed by more detailed examination for subtle abnormalities.

\begin{figure}[!t]
    \centering
    \includegraphics[width=0.4\linewidth]{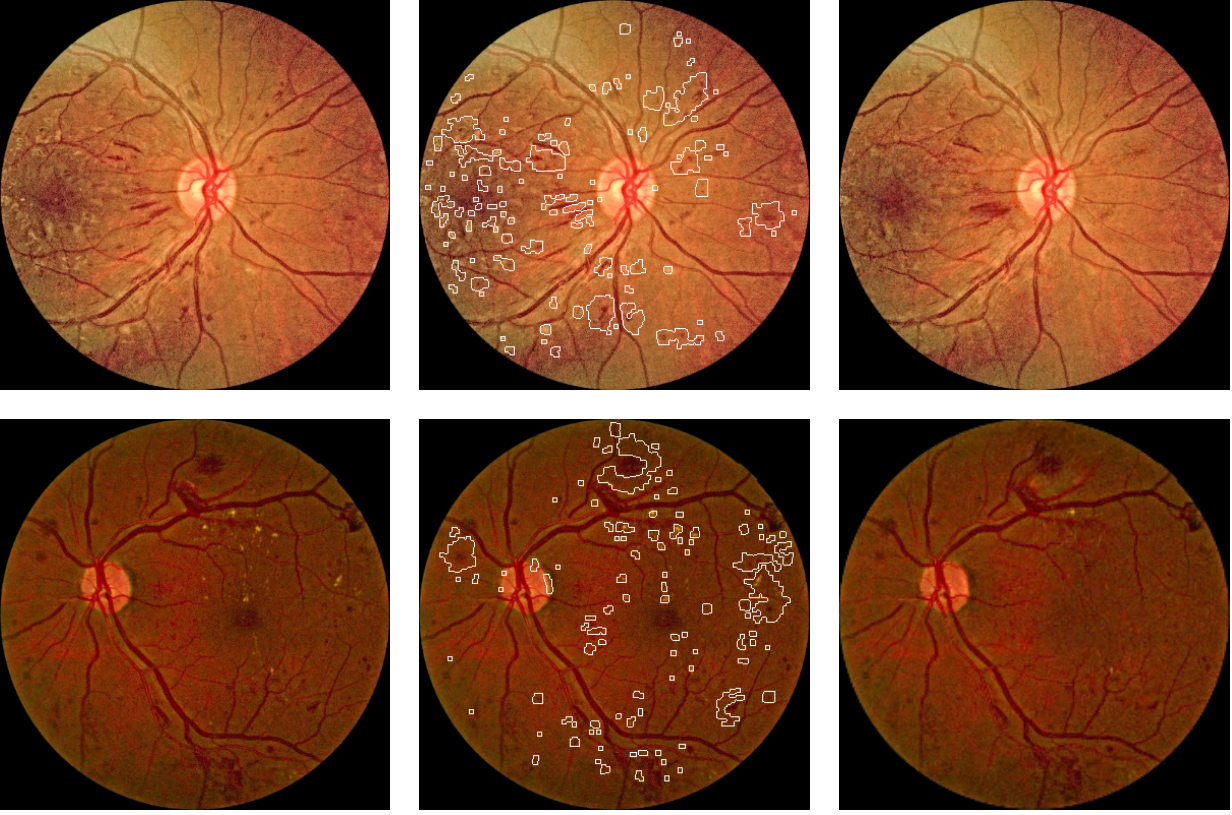}
    \caption{Failure cases inpainted or weakly semantic segmentation based on our method TWLR.}
    \label{fig:failure_case}
\end{figure}

\paragraph{Fully-inpainted fundus images Analysis}
Figure 6 demonstrates the clinical efficacy of our TWLR method through representative examples of successful DR remediation via computational lesion inpainting. The results showcase the method's ability to transform lesion-affected retinal images into healthy-appearing fundus photographs that achieve no DR classification. Each row presents a complete treatment pipeline: the left column displays original fundus images containing various DR lesions, the middle column illustrates the lesion classification accuracy with predicted regions (white) overlaid against ground truth annotations (green: hard exudates, purple: hemorrhages/exudates, light blue: microaneurysms, orange: soft exudates), and the right column shows the final inpainted images classified as no DR status.

The qualitative results reveal the method's proficiency in handling diverse lesion presentations across different retinal regions and lesion distributions. In the first case, scattered hard exudates and microaneurysms distributed throughout the macular region are successfully identified and remediated, resulting in a restored retinal appearance with preserved vascular architecture and natural tissue texture. The second case demonstrates effective treatment of more severe lesions, including extensive hard exudates and hemorrhages, while maintaining the integrity of optic disc characteristics and retinal vessel patterns. The third and fourth cases illustrate the method's capability to address complex lesion configurations, including clustered lesions and mixed lesion types, achieving seamless inpainting that preserves anatomical realism.
The preservation of retinal anatomical structures, including blood vessel continuity, optic disc morphology, and macular architecture, validates the sophistication of the inpainting algorithm. The seamless integration of remediated regions with surrounding healthy tissue demonstrates that the method successfully maintains photorealistic quality while eliminating lesion features. These qualitative outcomes support the quantitative performance metrics, confirming that the TWLR approach achieves meaningful DR grade reduction through targeted lesion removal while preserving the essential anatomical and visual characteristics of healthy retinal tissue. The consistent success across diverse lesion presentations underscores the clinical potential of computational DR remediation as a viable approach for achieving healthier retinal status classification.

\paragraph{Failure cases}
Figure \ref{fig:failure_case} presents representative failure cases encountered during inpainting or WSSS using our TWLR method, illustrating the inherent limitations and challenges in computational DR remediation. These cases provide critical insights into the boundary conditions where the current approach faces difficulties, offering valuable guidance for future methodological improvements.
The upper case demonstrates a challenging scenario involving extensive hard exudates distributed across the central macular region. Despite accurate lesion classification (middle panel), the inpainting process fails to achieve satisfactory restoration, resulting in visible artifacts or incomplete lesion removal in the final output (right panel). This failure likely stems from the high density of lesion features and their proximity to critical anatomical structures, where aggressive inpainting may compromise essential retinal details or create unrealistic tissue appearances.
The lower case illustrates another problematic scenario characterized by severe proliferative DR with extensive hemorrhages and exudates occupying a substantial portion of the fundus. The complex lesion architecture and the large area requiring restoration exceed the method's current capabilities, leading to suboptimal inpainting results. The failure in this case highlights the challenges associated with severe DR grades where lesion changes are too extensive to permit realistic restoration while maintaining anatomical integrity.
These failure cases underscore several key limitations of the current approach: (1) difficulty handling extensive lesion areas that compromise large retinal regions, (2) challenges in preserving fine anatomical details when lesions are located near critical structures like the fovea, and (3) limitations in addressing severe proliferative changes that fundamentally alter retinal architecture. The analysis of these failure modes is essential for understanding the method's operational boundaries and provides direction for future research focusing on advanced inpainting techniques capable of handling more complex lesion presentations while maintaining photorealistic quality and anatomical accuracy.

\section{ACKNOWLEDGMENTS}
This work is partially supported by Guangdong Provincial Key Laboratory of Interdisciplinary Research and Application for Data Science, Beijing Normal-Hong Kong Baptist University, project code 2022B1212010006, BNBU research grant R0400001-22; National Natural Science Foundation of China (No.12231004), BNBU research grant UICR0600048; National Natural Science Foundation of China (No.1272054), BNBU research grant UICR0600036; Guangdong University Innovation and Enhancement Programme Funds Featured Innovation Project 2018KTSCX278.

\bibliographystyle{unsrt}  
\bibliography{references}  

\end{document}